\pdfoutput=1

\documentclass[11pt]{article}

\usepackage[final]{acl}

\usepackage{times}
\usepackage{latexsym}

\usepackage{xcolor}
\definecolor{aquamarine}{rgb}{0.5, 1.0, 0.83}
\definecolor{ballblue}{rgb}{0.13, 0.67, 0.8}
\definecolor{brandeisblue}{rgb}{0.0, 0.44, 1.0}
\usepackage[T1]{fontenc}

\usepackage[utf8]{inputenc}

\usepackage{microtype}

\usepackage{inconsolata}

\usepackage{graphicx}
\usepackage{multirow}
\usepackage{amsmath}
\usepackage{amssymb}
\usepackage{float}
\usepackage{overpic}
\usepackage{subfig}
\usepackage{color}
\usepackage{url}
\usepackage{hyperref}
\title{Paying More Attention to Source Context: Mitigating Unfaithful Translations from Large Language Model}

\author{Hongbin Zhang$^{\dag\ddag}$,Kehai Chen$^{\dag}$\thanks{Corresponding Author},Xuefeng Bai$^{\dag}$,Yang Xiang$^{\ddag}$,Min Zhang$^{\dag}$ \\
\textsuperscript{\dag}Institute of Computing and Intelligence, Harbin Institute of Technology, Shenzhen, China \\
\textsuperscript{\ddag}Peng Cheng Laboratory, Shenzhen, China \\
\texttt{azure.starzhang@gmail.com,\{chenkehai,baixuefeng,zhangmin2021\}@hit.edu.cn,} \\
\texttt{xiangy@pcl.ac.cn} \\}

\begin{document}
\maketitle
\begin{abstract}
Large language models (LLMs) have showcased impressive multilingual machine translation ability. However, unlike encoder-decoder style models, decoder-only LLMs lack an explicit alignment between source and target contexts.
Analyzing contribution scores during generation processes revealed that LLMs can be biased towards previously generated tokens over corresponding source tokens, leading to unfaithful translations. 
To address this issue, we propose to encourage LLMs to pay more attention to the source context from both source and target perspectives in zeroshot prompting: 1) adjust source context attention weights; 2) suppress irrelevant target prefix influence; Additionally, we propose 3) avoiding over-reliance on the target prefix in instruction tuning.
Experimental results from both human-collected unfaithfulness test sets focusing on LLM-generated unfaithful translations and general test sets, verify our methods' effectiveness across multiple language pairs.
Further human evaluation shows our method's efficacy in reducing hallucinatory translations and facilitating faithful translation generation.
\footnote{The code and data are released on \url{https://github.com/AzureStarz/paying_attention_to_the_source.git}.} 
\end{abstract}

\section{Introduction}

Large language models (LLMs; \citealt{NEURIPS2020_1457c0d6,liu2023pre})
have shown great potential in machine translation within recent years~\citep{lin-etal-2022-shot,zhang2022opt,hendy2023good,jiao2023ischatgpt}. Given the different modeling architectures and pre-trained objectives in decoder-only LLMs and encoder-decoder neural machine translation models, previous studies have probed into leveraging LLMs for translation via in-context learning~\citep{zhu2023multilingual,vilar-etal-2023-prompting,zhang2023prompting} or instruction tuning~\citep{jiao-etal-2023-parrot,muennighoff-etal-2023-crosslingual,xu2023paradigm,alves2024tower}.

However, decoder-only LLMs, lacking an explicit mechanism, such as cross-attention modules~\citep{DBLP:journals/corr/BahdanauCB14,NIPS2017_3f5ee243} in encoder-decoder architectures, for aligning the source and target context. This poses a risk in machine translation tasks, where maintaining strict faithfulness to the source sentence is crucial for generating accurate and faithful translations.
\begin{figure}[!t]
    \centering
    \includegraphics[width=1\linewidth]{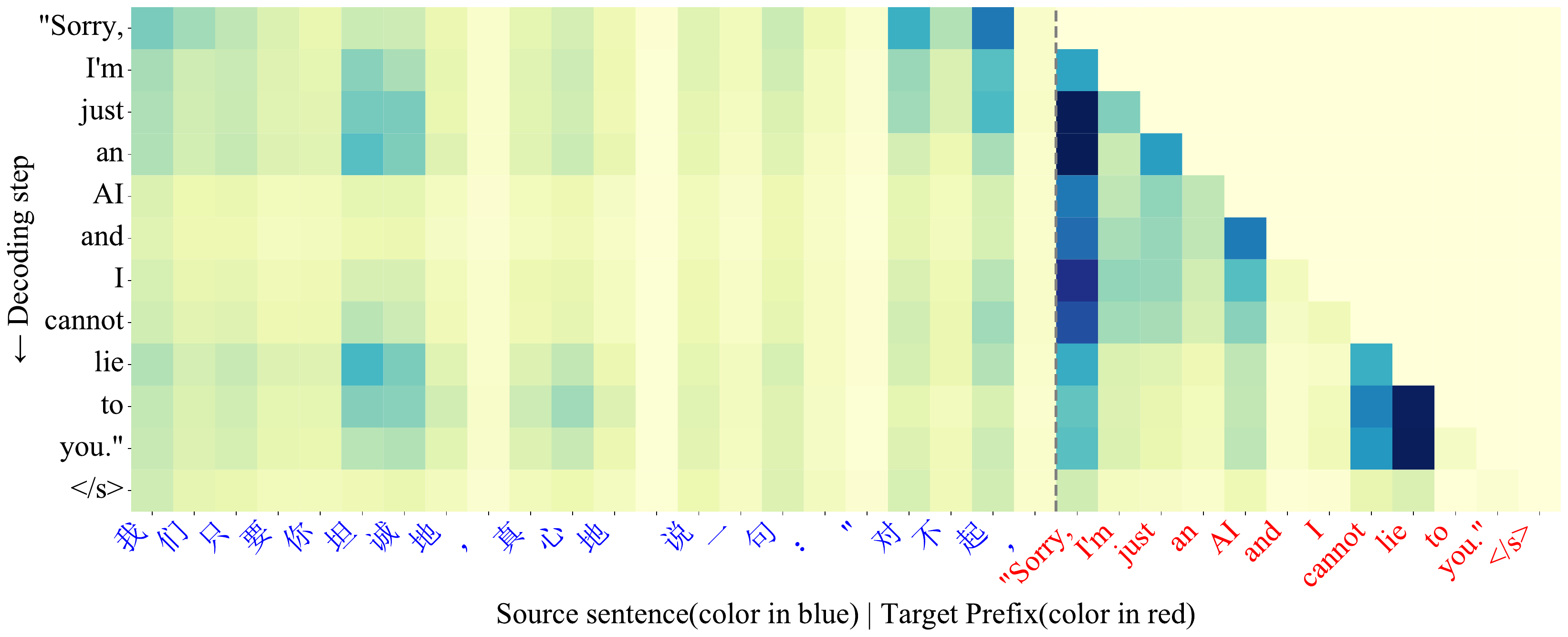}
    \caption{Contribution visualization of a Zh$\Rightarrow$En unfaithful translation instance. Each predicted token (row) corresponds to the contribution of each input token including source tokens and target prefixes (column) to the output token. One of the correct translations of the given source sentence is ``\textit{We just want you to honestly and sincerely say Sorry}''.}
    \label{intro_case}
\end{figure}
In Figure~\ref{intro_case}, we visualize the influence of the source and target tokens on the generating tokens during the generation process using contribution scores(developed from~\citet{ferrando-etal-2022-towards,ferrando-etal-2022-measuring} and be detailed in Apendix~\ref{alti_sec}) in decoder-only LLMs, e.g., Llama-2-7B. As shown, we observe two counter-intuitive phenomena: 1) LLMs pay much attention to the previously generated token ``\textit{Sorry}'' throughout almost the entire generation process and 2) they less focus on the source tokens corresponding to the generating target tokens. This leads to LLMs generating the hallucinatory response ``\textit{Sorry, I'm just an AI and I cannot lie to you.}'', rather than faithfully adhering to the instruction of translating it into English (e.g., ``\textit{We just want you to honestly and sincerely say Sorry}''). 

To tackle this issue, we propose strategies targeting both source and target aspects to guide the decoder-only LLMs toward focusing more on the source context during the generation process.
Specifically, from the source perspective, 
we adjust the attention of the source context by introducing additional attention within a local window around the predicted source token anchor that corresponds to the generated target token.
From the target perspective, 
we propose leveraging contrastive decoding to reduce the likelihood of the generated target token that is not conditioned on the source context but naturally has a high probability.
Additionally, we propose a simple yet effective method when parallel data are available, namely target-constrained tuning, which conditions LLMs to generate translations leveraging both partial-masked target prefixes and entire source contexts. Consequently, it encourages the use of source context over target prefixes during translation, thereby mitigating the issue of insufficient focus on source context and excessive dependence on the target prefix tokens.
We take LLaMA-2 series~\citep{touvron2023llama} as backbones and conduct experiments in both unfaithful translation test sets and open benchmarks, like WMT22~\citep{kocmi-etal-2022-findings} and Flores~\citep{flores101}.
Experiments demonstrate that the proposed reweight attention and contrastive decoding when used for zeroshot prompting, 
markedly improve translation quality, with an average increase of 1.7 BLEU and 4.0 COMET scores compared to vanilla prompting. 
Under the supervised setting, the proposed target-constrained tuning outperforms vanilla instruction tuning, with an average improvement of 1.1 in BLEU score and 0.6 in COMET score.
Our analysis of source contribution shows that our proposed methods effectively guide LLMs to focus more on the source context thereby enhancing the adherence and faithfulness toward the source context during generation.
Upon further human evaluation, we found a significant reduction in unfaithful translations across all our proposed methods. Our main contributions are summarized as follows:
\begin{itemize}
    \item This paper first focuses on the issue that the LLM-based MT over-depends on the generated target-side contextual information due to lacking cross-attention, which leads to more unfaithful translation.
    \item This paper proposes three methods aimed at different application scenarios
    to improve this serious phenomenon of unfaithful translation brought by the target-side context bias. 
    \item We annotate a specific unfaithful dataset tailored for LLMs to evaluate the effectiveness of the proposed approach. 
\end{itemize}

\section{Methodology}
\begin{figure*}[htbp]
	\centering
 	\subfloat[Paradigms of reweight attention and contrastive decoding]{\includegraphics[width=.5\linewidth]{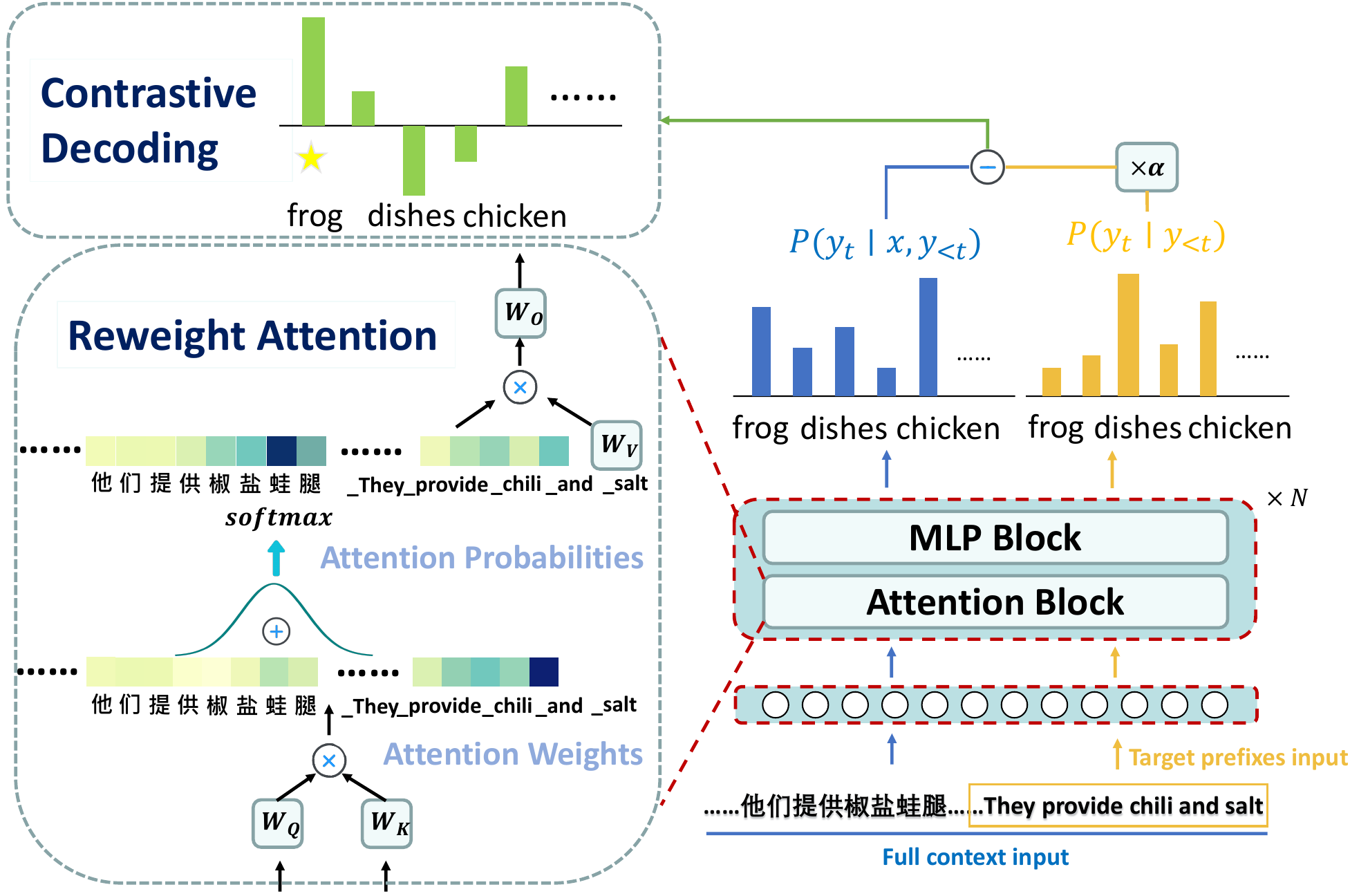}\label{unsupervised}}
	\subfloat[Framework of target-constrained tuning]{\includegraphics[width=.5\linewidth]{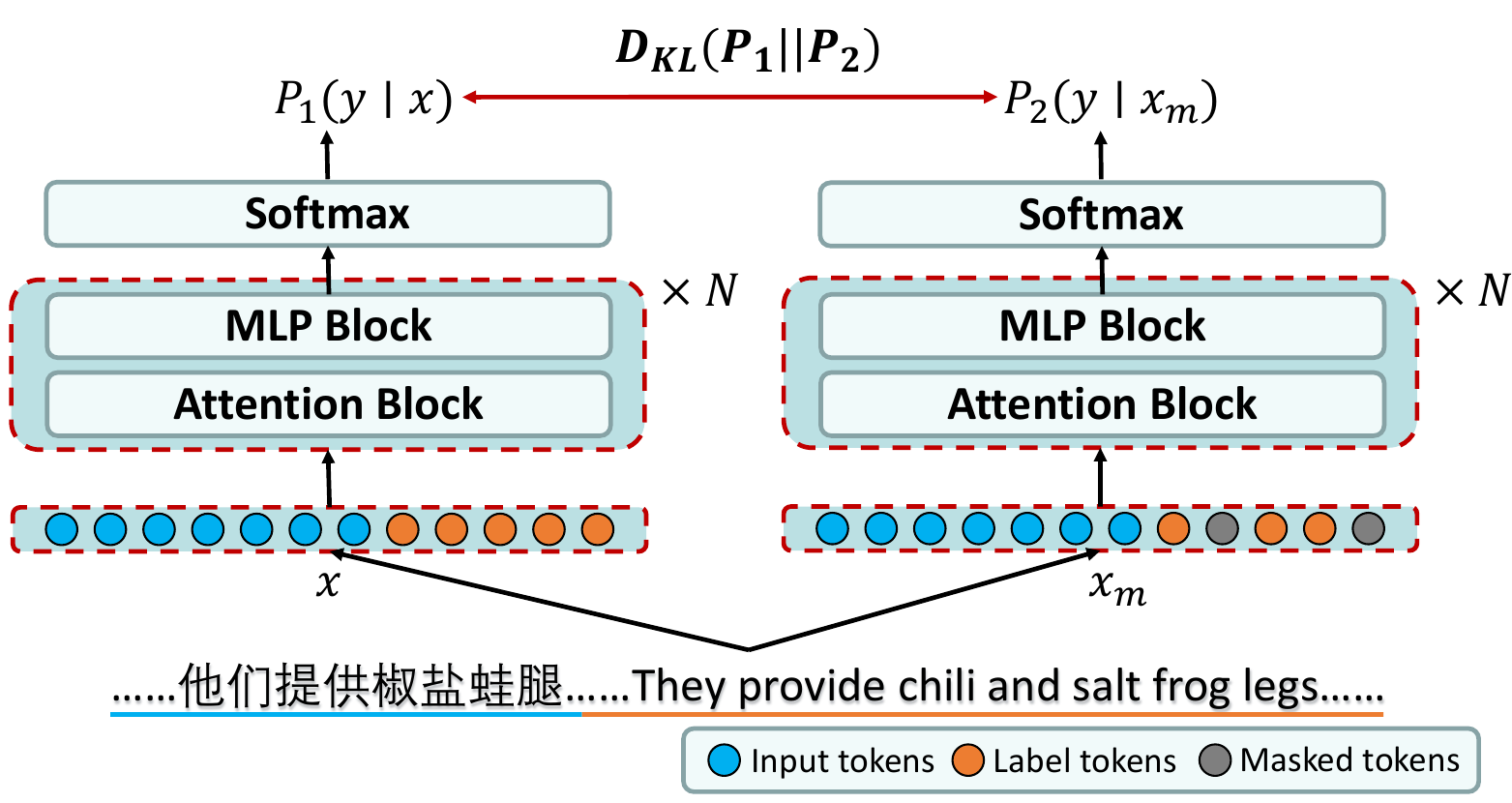}\label{tct}}
	\caption{The left picture shows the paradigms of proposed unsupervised methods, including the reweight attention and contrastive decoding. The right picture illustrates the target-constrained tuning, detailing how the two different inputs, full input $x$ and label-masked input $x_{m}$ will go through the model and obtain two distributions $P_1$ and $P_2$.}
    \label{proposed_methods}
\end{figure*}

Recognizing the significance of both source and target aspects in machine translation, we address the aforementioned issues by enhancing the contribution of the source context and diminishing the influence of the target prefixes. Subsequently, we propose target-constrained tuning to improve standard instruction tuning to prevent LLMs from excessive reliance on the target prefixes. The overview of our proposed methods is shown in Figure \ref{proposed_methods}.

\subsection{Boosting Source Influence: Reweight Attention}

Our reweight attention mechanism is based on a local window, drawing inspiration from \citet{luong-etal-2015-effective}. 
This mechanism selectively focuses on the subset of the source context during translation. 
In more detail, the model initially determines an aligned position $p_t$ for each target token at time $t$. The local attention weight is subsequently derived from query vectors and key vectors corresponding to the source context within the window $[p_t - D, p_t + D]$, where the value of $D$ is empirically determined. Subsequently, we explore two variants of the method, outlined below:
\paragraph{Monotonic Alignment:} We straightforwardly set $p_t = t$, assuming a rough monotonic alignment between source and target sequences.
\paragraph{Contribution-guided Alignment:} We propose leveraging the contribution measurement introduced by \citet{kobayashi-etal-2021-incorporating} to heuristically 
designate the most significant source token from the entire source context as the aligned position. 
\begin{equation}
    p_t =\underset{i} \max  ||(\text{LN}(\sum_{i=1}^{S}T_t(x_i) + b_{O} + x_t))||_2,
\end{equation}
Here, $T_t(x_i)$ represents the linear transformation detailed in Appendix \ref{alti_sec}, $\text{LN}$ represents the layer normalization operation, $b_O$ represents the bias term of the output linear projection, $x_t$ represents the residual connection and $S$ denotes the length of the source sentence.

To promote alignment points near $p_t$, we model a Gaussian distribution centered around $p_t$. Specifically, the alignment vectors $\alpha_t$ are defined as:
\begin{equation}
    \alpha_t(s) = \omega \times exp(-\frac{(s-p_t)^2}{2\sigma^2}),
\end{equation}
Here, $\omega$ serves as the scale factor regulating the additional attention weight, $s$ represents the index of the source tokens, and the standard deviation $\sigma$ is empirically set to $\frac{D}{2}$. Subsequently, we modify the attention output by adding extra attention weights calculated by the local attention window:
\begin{equation}
\resizebox{1\hsize}{!}{
$\text{attention}(Q_t,K_x,V_x) = \text{softmax}((1 + \alpha_t) \times \frac{Q_t K_x^T}{\sqrt{d_k}}) V_x$,
}
\end{equation}
where $Q_t=W_{Q}x_t, K_x=W_{K}x, V_x=W_{V}x$ denote the query vector of target position $t$, key vectors, and value vectors of input $x$, respectively. $d_k$ represents the dimension of a single vector. 

\subsection{Mitigating Target Impact: Contrastive Decoding}
To prevent undesired generations that are not conditioned on the source context,
we propose facilitating LLMs to diminish the contribution of the target prefix through contrastive decoding~\citep{li-etal-2016-diversity,2023arXiv230514739S}. 

By replacing the standard log-likelihood objective function with the maximum mutual information (MMI) as an alternative objective function $O$, we select tokens that maximize the mutual information between the input context $X$ and the translation output $Y$:
\begin{equation}
   O_{MMI} = \log \frac{P(X,Y)}{P(X)P(Y)},
\end{equation}
$P(\cdot)$ is estimated by providing the LLM with the translation instruction prompt as shown in Appendix~\ref{appendix_exp_setting}. This prevents bias towards translations that may inherently carry a high probability without being conditioned on the source context. Instead, it encourages responses that are specifically tailored to the given source input. 
Moreover, we extend the MMI objective which introduces a hyperparameter $\alpha$ that controls the degree to which unconditional responses are penalized:
\begin{equation}
\resizebox{1\hsize}{!}{
    $y_t = \underset{\nu}{\arg \max}\{\log p(y_t\mid x,y_{<t}) - \alpha \log p(y_t\mid y_{<t})\},$
}
\end{equation}
where $x$ is the input query, and $y_{<t}$ is the response before timestep $t$.

\subsection{Target-constrained Instruction Tuning}
\label{tct_sec}
Based on the previous analysis and inspired by \citet{NIPS2015_e995f98d} and \citet{NEURIPS2021_5a66b920}, we propose target-constrained instruction tuning to encourage LLMs to learn generating translations given the entire source context and incomplete target prefixes, thereby preventing over-relying on target prefixes when generation.

Concretely, given the instruction style query $x$ which contains the translation instruction and source sentence as constructed in Appendix~\ref{appendix_exp_setting}, and target sentence $y$ as the label for supervised training, we first feed the full instruction $\{x,y\}$ to go through the forward pass of the model to obtain the distribution of the model predictions denoted as $\mathcal{P}^{f}(y_t|x,y_{<t})$. We then generate the partially masked targets $y^{m}$, where target tokens are masked with a probability $\beta$. Following this, the target-constrained instruction input $\{x,y^{m}\}$ is fed into the model, resulting in a target-constrained distribution for the model's prediction, represented as $\mathcal{P}^{c}(y_t|x,y^{m}_{<t})$. 
During the training step, our method aims to regularize model predictions by minimizing the bidirectional Kullback-Leibler(KL) divergence between the output distributions corresponding to the same source context, which is:
\begin{equation}
\resizebox{1\hsize}{!}{$
\begin{split}
    \mathcal{L}_{KL} = \frac{1}{2} \Big( \mathcal{D}_{KL}\big(\mathcal{P}^{f}(y_{t} \mid x,y_{<t}) \parallel \mathcal{P}^{c}(y_{t} \mid x,y^{m}_{<t})\big) + \\ 
    \mathcal{D}_{KL}\big(\mathcal{P}^{c}(y_{t} \mid x,y^{m}_{<t}) \parallel \mathcal{P}^{f}(y_{t} \mid x,y_{<t})\big) \Big).
\end{split}$}
\end{equation}
Building upon the basic negative log-likelihood learning objective $\mathcal{L}_{NLL}$ associated with the two forward passes:
\begin{equation}
\small
\mathcal{L}_{NLL}=-\log \mathcal{P}^{f}\left(y_{t} \mid x,y_{<t}\right)-\log \mathcal{P}^{c}\left(y_{t} \mid x,y^{m}_{<t}\right),
\end{equation}
To sum up, we jointly optimize the total loss, incorporating full context translation loss, target-constrained translation loss, and regularized KL-Divergence loss, as illustrated below:
\begin{equation}
    \mathcal{L} = \mathcal{L}_{NLL} + \lambda \cdot \mathcal{L}_{K L},
\end{equation}
where $\lambda$ is the coefficient weight to control $\mathcal{L}_{K L}$.
By minimizing this loss, 
the probability distribution of the entire input context becomes less dependent on the target prefixes, thereby encouraging LLMs to utilize the source context to the greatest extent for generating translation.

\section{Experiments}

\begin{table*}[!ht]
\centering
\resizebox{\linewidth}{!}{\begin{tabular}{lcccccccc}
\hline
\multirow{2}{*}{System} &
  \multicolumn{2}{c}{De $\Rightarrow$ En} &
  \multicolumn{2}{c}{En $\Rightarrow$ De} &
  \multicolumn{2}{c}{Zh $\Rightarrow$ En} &
  \multicolumn{2}{c}{En $\Rightarrow$ Zh} \\ \cline{2-9} 
                          & BLEU & COMET & BLEU & COMET & BLEU & COMET & BLEU & COMET \\ \hline
\multicolumn{9}{l}{\textbf{Unsupervised Setting}}                                             \\
Vanilla Zeroshot          & 23.2  & 77.8   & 7.11  & 60.6   & 11.4  & 74.3   & 3.81  & 42.8   \\
Reweight Attention (RA)    & \textbf{24.5}  & \textbf{79.0}   & \textbf{10.6}  & \textbf{63.3}   & \textbf{12.5}  & \textbf{75.0}   & \textbf{6.14}  & \textbf{57.0}   \\
Contrastive Decoding (CD)  & 24.2  & 78.7   & 9.10  & 61.5   & 12.1  & 74.6   & 4.81  & 53.7   \\
\hline
\multicolumn{9}{l}{\textbf{Supervised Setting}}  \\
Vanilla Fewshot           & 27.1  & 81.7   & 15.7  & 74.2   & 14.2  & 76.5   & 16.2  & 73.1   \\
Vanilla Instruction Tuning& 27.3  & 82.2   & 19.8  & 77.5   & 15.7  & 76.4   & 17.5  & 74.8   \\
Target-constrained tuning & 29.6  & 83.0   & 20.9  & 78.0   & 16.1  & 77.0   & 17.9  & 75.2   \\
Vanilla Instruction tuning LoRA              & 29.1  & 82.9   & 20.3  & 78.5   & 15.5  & 76.5   & 18.6  & 76.2   \\
Scheduled Sampling tuning LoRA              & 30.5  & 83.1   & 20.2  & 78.5   & 16.5  & 76.8   & 18.6  & 76.1   \\
R-Drop tuning LoRA              & 30.3  & 83.1   & 20.1  & \textbf{78.8}   & \textbf{16.6}  & \textbf{77.1}   & 18.7  & \textbf{76.5}   \\
Target-constrained tuning LoRA & \textbf{30.8}  & \textbf{83.2}   & \textbf{20.6}  & 78.5   & \textbf{16.6}  & 77.0   & \textbf{19.1}  & \textbf{76.5}   \\
\hline
\end{tabular}}
\caption{Translation performance of LLaMA2-7b-chat model on human-collected unfaithful translation test sets. The bold number marks the best metric results from the methods under the same translation evaluation setting.}
\label{main_result}
\end{table*}

We conduct experiments on the proposed human-collected unfaithful translation test sets containing unfaithful translations
covering three languages and four translation directions. Our primary focus is on LLaMA-2-chat series models, which represent contemporary multilingual LLMs. More details of experimental settings can be found in Appendix \ref{appendix_exp_setting}. The ablation study of the proposed methods can be referred to Appendix \ref{appendix_ablation}.
\subsection{Experimental Settings}
\paragraph{Dataset} 
We heuristically gather translation data that is prone to be unfaithful or hallucinatory based on the metric detailed in the Appendix~\ref{alti_sec} for all four translation directions(Chinese$\Leftrightarrow$English and German$\Leftrightarrow$English) as the evaluation data. We utilize human-written data from past WMT competitions rather than public training data to prevent the introduction of noises into instruction tuning following~\citet{jiao-etal-2023-parrot} and \citet{xu2023paradigm}. 
We employ newstest2017-2021 of Chinese$\Leftrightarrow$English and newstest2014-2021 of German$\Leftrightarrow$English tasks~\citep{post-2018-call},
This yields a total of 62.9K training sentence pairs data for all four directions. 

\paragraph{Baseline}
We train the model in a bilingual translation manner separately for different translation directions and use LLaMA-2-7B-chat as our backbone model given its best zero-shot and instruction following performance. 

\textbf{Vanilla Instruction Tuning/Vanilla Instruction Tuning LoRA} \textit{Full-Weight} or \textit{LoRA} vanilla instruction tuning on high-quality parallel data for LLaMA-2-7B-chat.

\textbf{Scheduled Sampling Tuning LoRA/R-Drop Tuning LoRA} \textit{LoRA} instruction tuning using Schedule Sampling~\citep{NIPS2015_e995f98d} or R-Drop~\citep{NEURIPS2021_5a66b920} on high-quality parallel data for LLaMA-2-7B-chat.

\paragraph{Metics.} We use BLEU~\citep{papineni-etal-2002-bleu} implemented in SacreBLEU\footnote{https://github.com/mjpost/sacrebleu}~\citep{post-2018-call}, and COMET\footnote{https://github.com/Unbabel/COMET}~\citep{rei-etal-2020-comet} from \textit{Unbabel/wmt22-comet-da}\footnote{https://huggingface.co/Unbabel/wmt22-comet-da} for automatic evaluation. 

\subsection{Main Results}
The results in Table \ref{main_result} reveal that the zeroshot prompting of the LLaMA2-7b-chat model exhibits poor performance on the unfaithful translation dataset. 
Comparing the baseline results with the improved outcomes achieved by our proposed methods across various translation directions and languages, we noted the following:
\paragraph{Elevating source focus brought improved translation quality.}
The reweight attention method outperforms vanilla zeroshot prompting, showing an average improvement of 2.1 BLEU and 4.7 COMET. It also exhibits superior performance in translations from English compared to translations to English. 
This observation might be attributed to the more severe inability to pay sufficient attention to specific source sentences when translating to languages other than English, leading to a decline in translation performance. 
The results suggest that the proposed reweight attention can enhance translation quality by directing LLaMA to prioritize the aligned source context.
\paragraph{Mitigation of target influence generates better translation.}
The contrastive decoding strategy significantly improves the translation performance of LLMs, outperforming the baseline with an average improvement of 1.2 BLEU and 3.3 COMET. However, the extent of improvement varies across different translation directions. It is noteworthy that contrastive decoding markedly enhances translations between German and English, while it results in only a marginal improvement in translations between Chinese and English compared to vanilla prompting. 
\paragraph{Target-constrained tuning refines vanilla instruction tuning.}
Instruction tuning only slightly improves translation performance compared to fewshot prompting, suggesting its limited effectiveness in addressing insufficient focus on the source context. However, the proposed target-constrained tuning consistently outperforms vanilla instruction tuning, with an average gain of 1.05 BLEU and 0.58 COMET. 
Furthermore, we employ low-rank adaptation~\citep{hu2022lora} to fine-tune the partial parameters of LLMs, aiming for improved efficiency. Experimental results demonstrate that LoRA tuning enhances performance across all translation directions and outperforms all parameters tuning. This is likely attributed to the limited tunable parameters in LoRA, preventing LLMs from overfitting to the small translation dataset and enhancing generalization ability~\citep{jiao-etal-2023-parrot}.
The results of the two baselines show that while both the Scheduled Sampling~\citep{NIPS2015_e995f98d} and R-drop~\citep{NEURIPS2021_5a66b920} methods can enhance performance to some degree, they are not as effective in mitigating unfaithful translations as our proposed target-constrained method.

\subsection{Human Evaluation}
Despite the utility of automatic evaluation metrics, they do not explicitly measure the degree to which our proposed method has mitigated the existence of unfaithful or hallucinated spans. Therefore, we conducted a human evaluation. Moreover, the consistency of hallucination evaluation between human evaluation and automatic metrics is studied in Appendix~\ref{connection_human_eval_automatic_metrics}.
\begin{figure}[!t]
    \centering
    \includegraphics[width=1\linewidth]{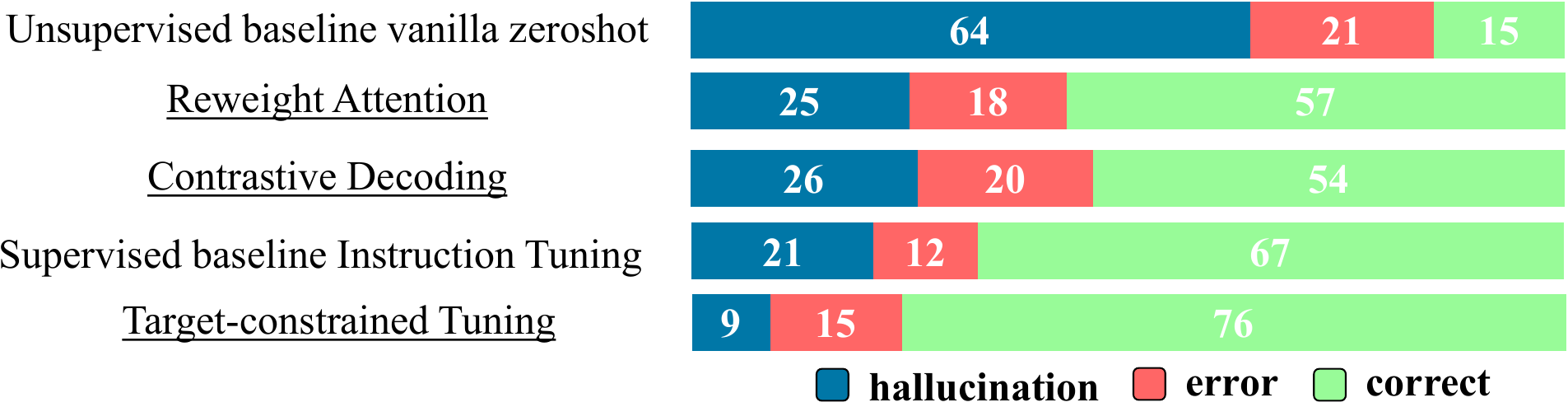}
    \caption{Human annotation results: percentages of translation categories for different methods.}
    \label{human_eval}
\end{figure}

\paragraph{Data.} 
Initially, we collected 100 instances, selected from the instances exhibiting the lowest cross-lingual sentence similarity between hypotheses and source sentences~\citep{dale-etal-2023-detecting,feng-etal-2022-language}. 
These sentences are likely to have hallucinated spans in the unfaithful translation dataset and were used as the human evaluation set. 
In total, 64\% of the sentences had their original translations marked as translation hallucination, 21\% as translation error, and 15\% as correct translations. 
Next, we use each method to translate the same set of 100 source sentences. The resultant sentence pairs are categorized by three annotators into three groups: Correct, Error, and Hallucination. 
More details on the annotation guidelines and inter-annotator agreement can be found in Appendix \ref{appendix_human_eval}.
\paragraph{Results.} The human evaluation results are displayed in Figure \ref{human_eval}. 
All proposed methods decrease the translation hallucination rate by at least a factor of 2.5. 
The unsupervised methods, including the reweight attention and contrastive decoding, significantly reduce hallucination in translations compared to baseline vanilla zeroshot. 
Interestingly, in terms of mitigating translation hallucinations, both the unsupervised methods perform on par with supervised instruction tuning.
However, the baseline instruction tuning exhibits fewer errors, which is anticipated, given that it was trained on a parallel corpus to generate accurate translations. 
When it comes to target-constrained tuning, we observe that it produces significantly fewer translation hallucinations than vanilla instruction tuning, but it results in more erroneous translations. Our analysis of the annotations reveals that while target-constrained tuning is capable of generating source-related translations, it also introduces some ambiguity issues, resulting in more errors. 
Overall, the application of all our proposed methods significantly reduces translation hallucinations.


\subsection{Generalize to other settings}
\begin{table}[!ht]
\centering
\resizebox{\linewidth}{!}{\begin{tabular}{ccccc}
\hline
\multirow{2}{*}{\begin{tabular}[c]{@{}c@{}}System\\ Base Model:LLaMA2-7b-Chat\end{tabular}} &
  \multicolumn{2}{c}{En $\Rightarrow$ De} &
  \multicolumn{2}{c}{Zh $\Rightarrow$ En} \\ \cline{2-5} 
                          & BLEU & COMET & BLEU & COMET \\ \hline
\multicolumn{5}{c}{Flores 101 Test Set}                 \\ 
Vanilla Zeroshot          & 21.8  & 78.7   & 19.3  & 83.9   \\
Reweight Attention        & 21.9  & 78.7   & 19.8  & 84.0   \\
Contrastive Decoding      & 21.7  & 78.5   & 19.5  & 84.1   \\ \hline
Vanilla Instruction tuning LoRA              & 29.2  & 84.6   & 22.9  & 84.6   \\
Target-constrained tuning LoRA & 29.5  & 84.7   & 24.5  & 85.1   \\ \hline
\multicolumn{5}{c}{WMT 22 Test Set}                         \\
Vanilla Zeroshot          & 23.5  & 74.4   & 19.7  & 77.4   \\
Reweight Attention        & 23.7  & 74.5   & 20.1  & 77.4   \\
Contrastive Decoding      & 23.4  & 74.4   & 19.7  & 77.5   \\ \hline
Vanilla Instruction tuning LoRA              & 34.0  & 81.7   & 24.4  & 77.6   \\
Target-constrained tuning LoRA & 35.3  & 82.1   & 25.3  & 78.5   \\ \hline
\end{tabular}}
\caption{Translation performance of LLaMA2-7b-chat model on Flores101 and WMT22 test sets}
\label{general_result}
\end{table}

\begin{figure*}[htbp]
	\centering
 	\subfloat[Contribution of the source context at each generation step]{\includegraphics[width=.5\linewidth]{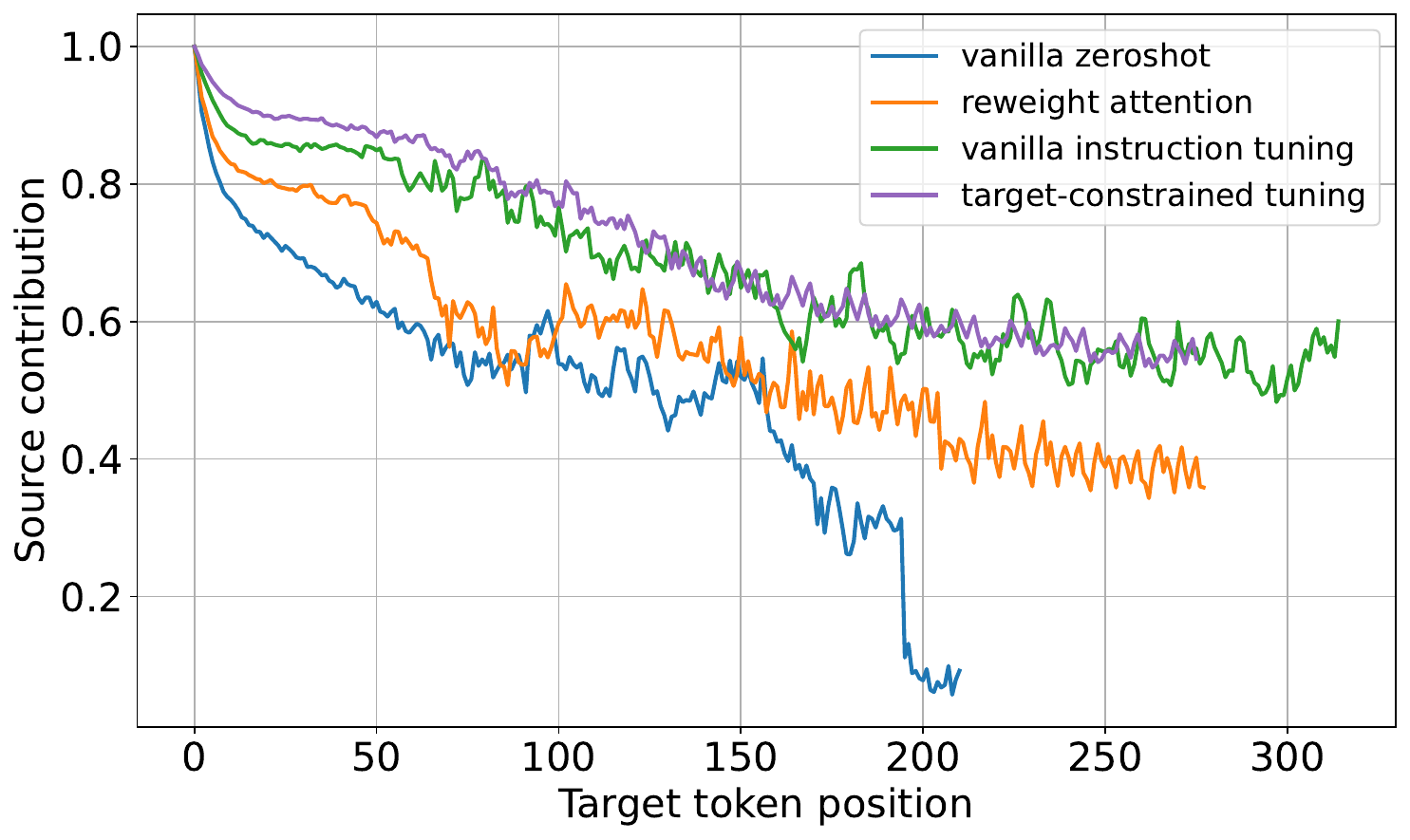}\label{contrib_comparison}}
	\subfloat[Entropy of the source contribution]{\includegraphics[width=.5\linewidth]{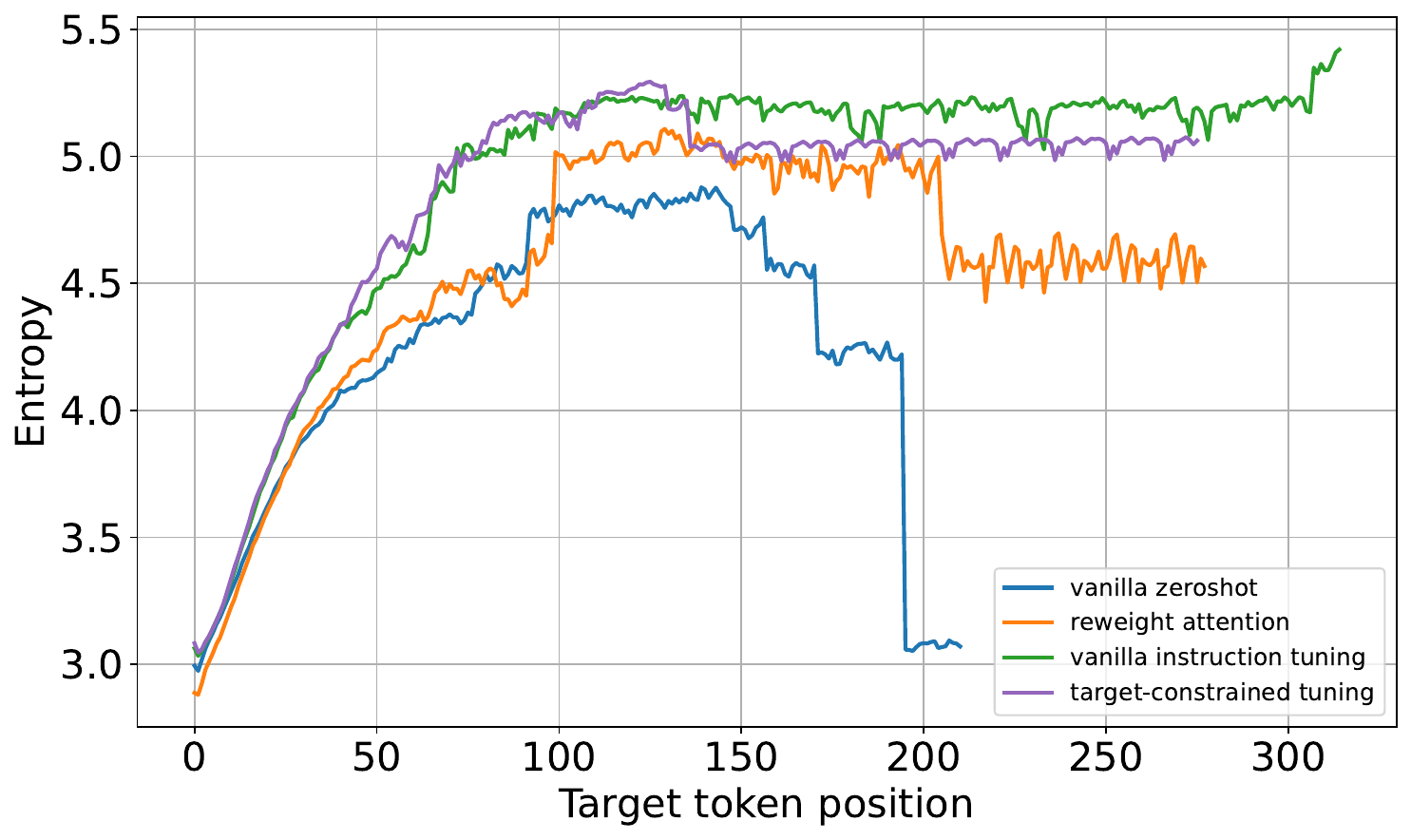}\label{contrib_src_entropy}}
	\caption{For each generation step, the figure shows the (a) contribution and (b) entropy of source context in the translation direction from Chinese to English. The points on the lines denotes the average score across the $i$-th target token position. Note that different methods result in different target generation lengths.}
    \label{contribution_distribution}
\end{figure*}

\paragraph{On general open dataset.}
While we have previously tested our methods on the proposed unfaithful translation dataset, we aim to determine their effectiveness on general open MT benchmarks. These include test sets such as Flores~\citep{flores101} and WMT22~\citep{kocmi-etal-2022-findings}.
As shown in Table \ref{general_result}, 
our proposed methods including the reweight attention and contrastive decoding still yield results comparable to vanilla zeroshot prompting. 
Target-constrained tuning continues to be effective in enhancing performance, given its dual function as a strategy to prevent overfitting on smaller translation datasets. 
Although there is an imbalanced improvement between the unfaithful test set and general test sets, our proposed methods can still achieve comparable performance on less hallucinated translation test sets. For a more in-depth analysis of this phenomenon, please refer to Appendix~\ref{analysis_imbalanced}.

\paragraph{Employing LLMs with different families and scales.}
\begin{table}[!t]
\centering
\resizebox{\linewidth}{!}{\begin{tabular}{ccccc}
\hline
\multirow{2}{*}{System} & \multicolumn{2}{c}{En $\Rightarrow$ De} & \multicolumn{2}{c}{Zh $\Rightarrow$ En} \\ \cline{2-5} 
                                & BLEU & COMET & BLEU & COMET \\ \hline
\multicolumn{5}{c}{Base Model:BLOOMZ-7b1-mt}                  \\
Vanilla Instruction tuning LoRA & 20.5 & 64.2  & 25.5 & 79.1  \\
Target-constrained tuning LoRA  & 21.2 & 64.4  & 26.2 & 79.4  \\ \hline
\multicolumn{5}{c}{Base Model:ChatGLM3-6b}                    \\
Vanilla Instruction tuning LoRA & 27.1 & 73.2  & 26.8 & 79.1  \\
Target-constrained tuning LoRA  & 27.2 & 73.2  & 27.2 & 79.2  \\ \hline
\multicolumn{5}{c}{Base Model:Vicuna-7b}                      \\
Vanilla Instruction tuning LoRA & 32.5 & 80.6  & 25.0 & 76.8  \\
Target-constrained tuning LoRA  & 33.5 & 81.3  & 25.2 & 76.9  \\ \hline
\end{tabular}}
\caption{Translation performance of various families of
LLMs with a similar size on WMT22 test sets.}
\label{model_families_wmt22}
\end{table}
\begin{table}[!t]
\centering
\resizebox{\linewidth}{!}{\begin{tabular}{ccccc}
\hline
\multirow{2}{*}{System} & \multicolumn{2}{c}{En $\Rightarrow$ De} & \multicolumn{2}{c}{Zh $\Rightarrow$ En} \\ \cline{2-5} 
                                & BLEU & COMET & BLEU & COMET \\ \hline
\multicolumn{5}{c}{Base Model:LLaMA2-7b-chat}                 \\
Vanilla Instruction tuning LoRA & 34.0 & 81.7  & 25.5 & 78.3  \\
Target-constrained tuning LoRA  & 34.5 & 81.8  & 25.8 & 78.4  \\ \hline
\multicolumn{5}{c}{Base Model:LLaMA2-13b-chat}                \\
Vanilla Instruction tuning LoRA & 37.0 & 83.2  & 27.6 & 79.3  \\
Target-constrained tuning LoRA  & 37.6 & 83.4  & 27.8 & 79.4  \\ \hline
\multicolumn{5}{c}{Base Model:LLaMA2-70b-chat}                \\
Vanilla Instruction tuning LoRA & 41.3 & 84.8  & 30.2 & 80.5  \\
Target-constrained tuning LoRA  & 41.8 & 85.0  & 31.1 & 80.8  \\ \hline
\end{tabular}}
\caption{Translation performance of various families of
LLMs with a similar size on WMT22 test sets.}
\label{model_size_wmt22}
\end{table}
We subsequently apply our target-constrained tuning to various LLM families that are of a similar size to LLaMA2-7b and evaluated on the WMT22 test sets, as shown in Table \ref{model_families_wmt22}. 
Despite the variations in architecture and pretraining corpus among these models, our target-constrained tuning method proves to be universally effective. 
In considering the different scales of LLMs, we select various scales of LLaMA2-chat versions to compare their performance on the WMT22 test sets, as depicted in Table \ref{model_size_wmt22}. Our proposed target-constrained tuning notably outperforms the vanilla instruction tuning. We have also conducted additional experiments on the unfaithful test set, with details provided in Appendix~\ref{more_comparison_exps}.

\section{Analysis}
\subsection{Contribution Distribution}
In this section, we reveal the contribution characteristic of input tokens' to generation observed in the behavior of the LLaMA2-7b-chat model. 
Specifically, during the translation from Chinese to English using our proposed unfaithful dataset, we compute the average contribution of source context generation and the entropy of the source contribution from each target token position. 
\paragraph{Changes of source contribution during generation.}
Following section \ref{alti_sec}, we can derive the input token contribution matrix $C(t,i)$, which denotes the contribution score of the $i_{th}$ input token during the generation of the $t_{th}$ target token. Specifically, for each generation step $t$, we compute the aggregate contribution from the source as $C_t(source) = \sum_{i \in S}C(t,i)$, where $S$ is the set of the indices corresponding to source tokens. 
As illustrated in Figure \ref{contrib_comparison}, we note that, during the entire generation process, the impact of the source diminishes (or, conversely, the impact of the prefix intensifies). This is an anticipated outcome: as the prefix lengthens, the model faces less uncertainty in determining which source tokens to utilize but needs to exert more control over fluency. These observations align with findings from previous work by \citet{voita-etal-2021-analyzing}.
It is evident that during the early phases of token generation, the source contributions of the reweight attention method surpass those of the vanilla zeroshot, and the source contributions of the target-constrained tuning exceed those of the baseline instruction tuning. 
Furthermore, there's a steep decline observed in the baseline zeroshot, further detailed in Appendix~\ref{analysis_sudden_drop}. In contrast, our methodologies show a gradual decline.
This suggests that our proposed methodologies can effectively amplify the significance of the source, and it also indicates the effectiveness of our methods in mitigating the issue associated with inadequate attention to the source context. 

\begin{figure*}[htbp]
	\centering
	\subfloat[Vanilla zero-shot prompting over-translation case.]{\includegraphics[width=.5\linewidth]{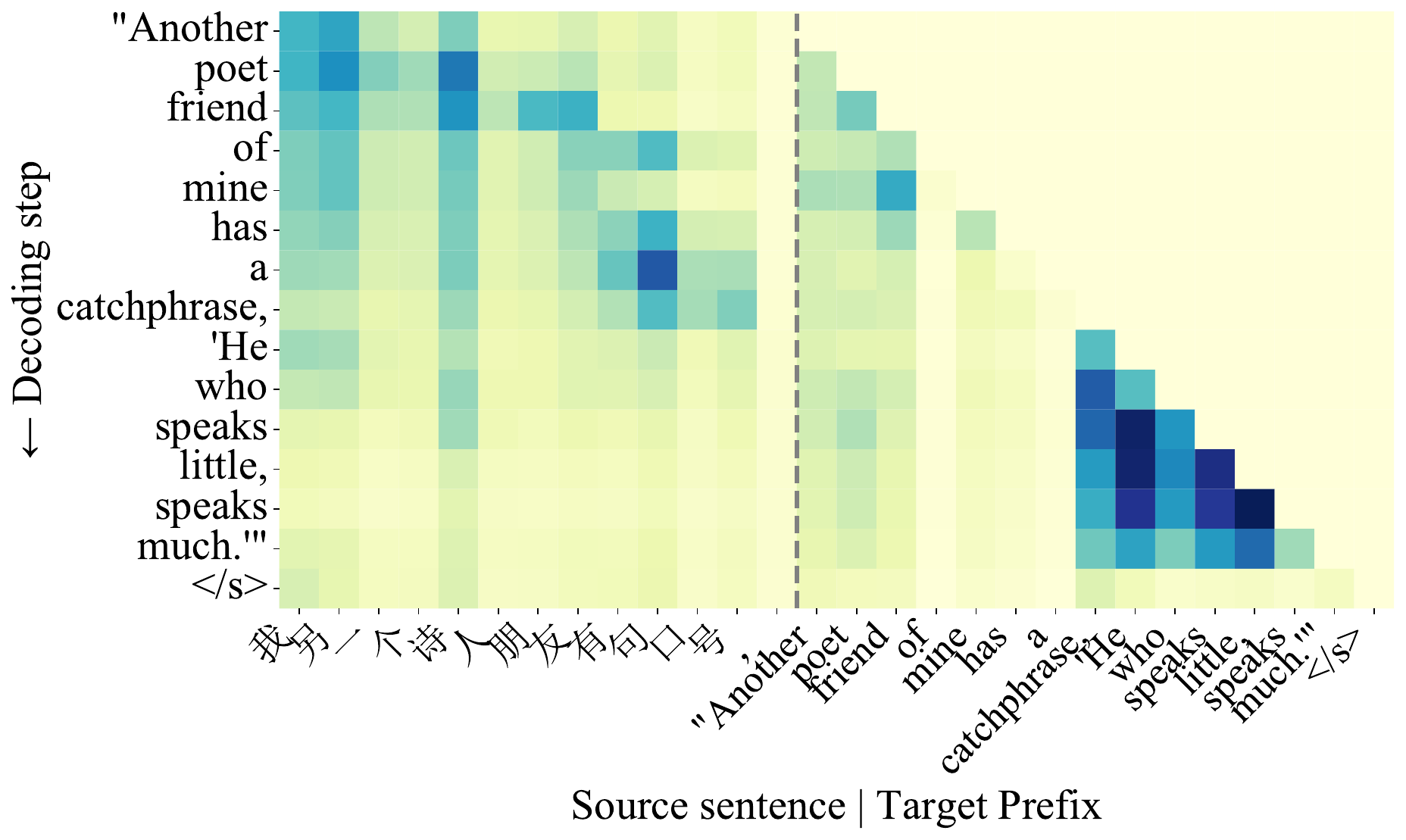}\label{zeroshot_hallu}}
	\subfloat[Reweight attention mitigate translation hallucination issue]{\includegraphics[width=.5\linewidth]{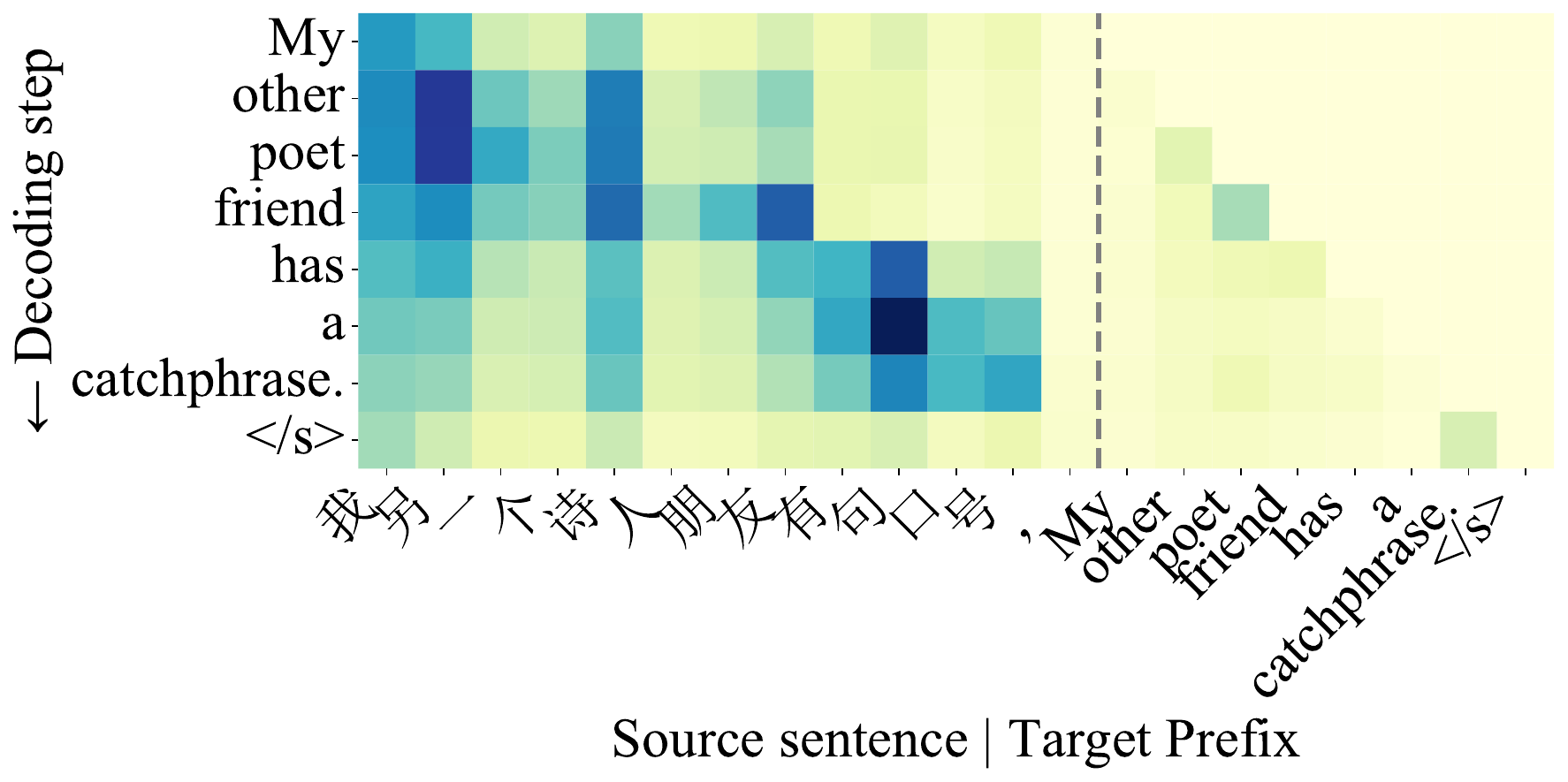}\label{ra_case_study}}\\
	\subfloat[Vanilla instruction tuning under-generation case]{\includegraphics[width=.5\linewidth]{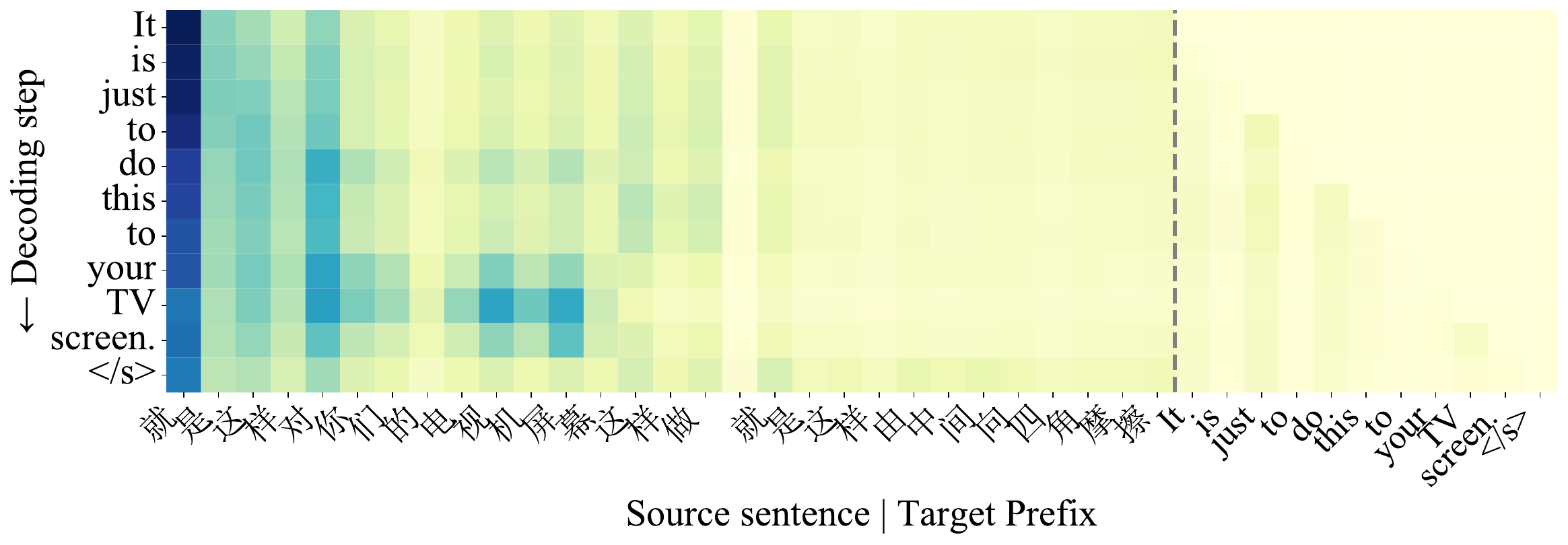}\label{van_lora_hallu}}
	\subfloat[Target-constrained tuning address omission issue]{\includegraphics[width=.5\linewidth]{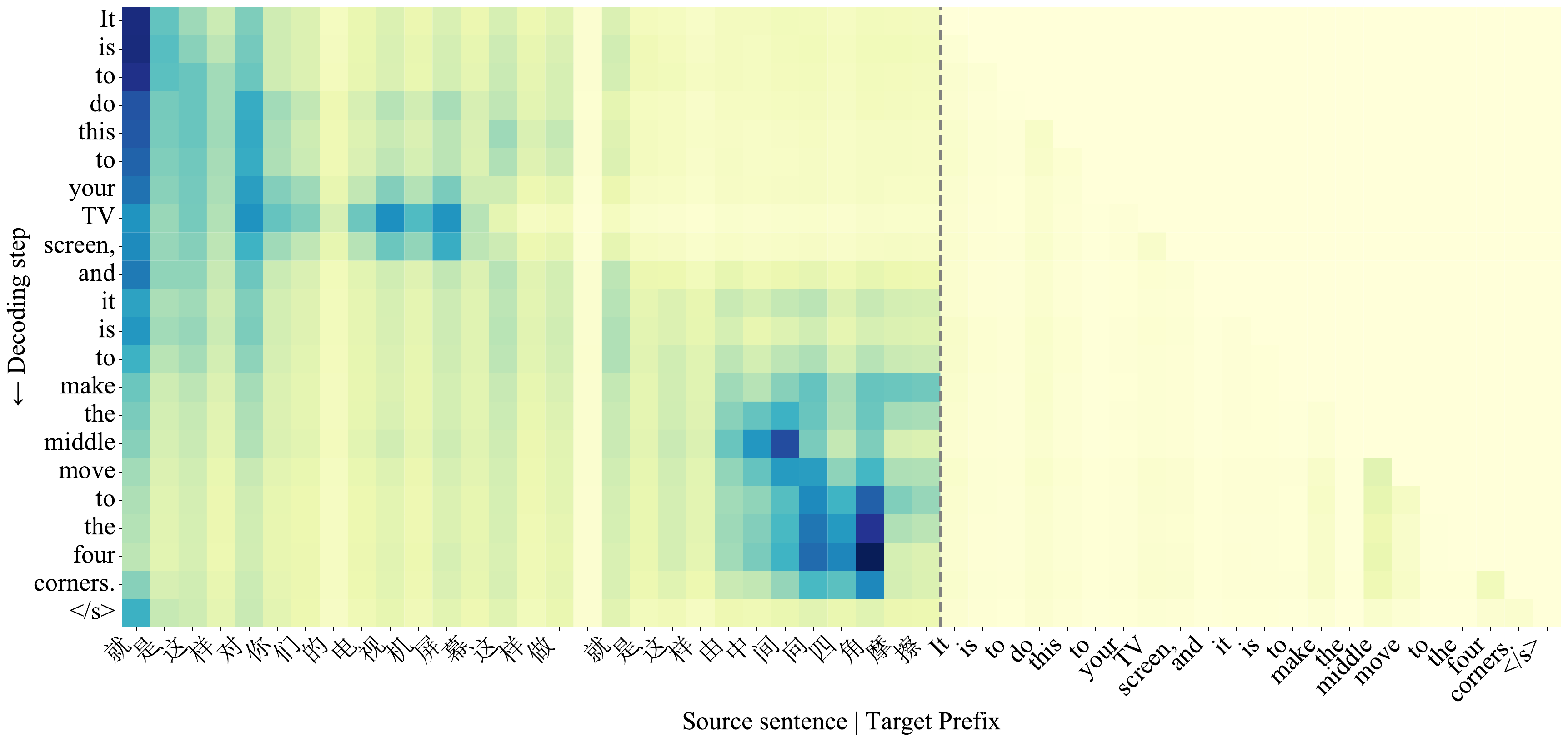}\label{tct_case_study}}
	\caption{Contribution visualization for unfaithful translations and corresponding mitigation across various settings. One of the correct translations of the first row is: ``\textit{Another poet friend of mine has a slogan.}''; One of the correct translations of the second row is: ``\textit{Just do this to your TV screen, rub it from the center to the four corners.}''}
    \label{case_study}
\end{figure*}

\paragraph{Entropy of source contributions.}
Let's examine the `sharpness' of the contributions of source tokens at different steps of generation. 
For each step, denoted by $t$, we calculate the entropy of the normalized source contributions, represented by $\{\frac{C(t,i)}{C_t(source)} \}_{i=1}^{S}$.
As shown in Figure \ref{contrib_src_entropy}, during the zeroshot generation phase of LLMs, we observe an increase in entropy until about the 150th position in the generated translation. Beyond this point, the entropy starts to diminish as the rest of the translation is generated. However, the baseline instruction tuning as well as our proposed methods invert this trend. After the generation of about the 150-th token, the entropy of source contributions remains high and fluctuates around 5. This indicates that when confronted with a longer context, the generation of LLMs necessitates a broader source context. Furthermore, our proposed methods increase the entropy of source contributions compared to the vanilla methods during the initial stage of translation generation. 
This implies that the application of our methods in generating translations requires a more comprehensive view of the source context, thereby enhancing the significance of the contribution from the source context.

\subsection{Case Study Using ALTI+}

For qualitative analysis, we present several hallucinated examples under different settings and corresponding mitigation instances from Chinese to English (Zh $\Rightarrow$ En) translation direction in Figure \ref{case_study}.
As expected, hallucination spans ought to be discernible in our contribution visualization method, either as an emphasis on specific target prefixes or as a drop in the contribution of the source sentence, as manifested in Figure \ref{zeroshot_hallu} and Figure \ref{van_lora_hallu}. 
Compared to the vanilla zeroshot baseline, the reweight attention approach can accurately direct the model's attention to the source context and generate source-aligned translation, thereby mitigating the hallucination issue as depicted in Figure \ref{ra_case_study}. 
It's crucial to emphasize that our reweight attention strategy doesn't impact the segments that are already accurately translated. It enhances the translation process when LLMs find it challenging to confidently pay attention to the source context by facilitating concentration on the relevant source tokens while generating the translation. 
When it comes to vanilla instruction tuning, the phenomenon of omissions frequently occurs in its generation(e.g., in Figure \ref{van_lora_hallu}), indicating that LLMs still struggle to cover the entire source context in translations, even after supervised tuning. By implementing target-constrained tuning, we ensure that LLMs do not excessively depend on the target prefix, but rather exploit as much of the source context as possible. This approach fosters the generation of more accurate and source-related translations than vanilla instruction tuning, as demonstrated in Figure \ref{tct_case_study}. 

\section{Related Works}
\subsection{Improving coverage of neural sequence to sequence model}
Over the past few years, the coverage for attention mechanism and hallucinations in neural machine translation(NMT) have been the subject of study for several years~\citep{tu-etal-2016-modeling,shan2021modeling,li2022faithfulness,lee2019hallucinations,muller-etal-2020-domain,dale-etal-2023-detecting}. \citet{tu-etal-2016-modeling} suggested the inclusion of coverage information to provide additional data about the probability of source words. \citet{mi2016coverage} introduced explicit coverage embedding models to mitigate issues of unfaithful translation in NMT. 
\citet{tu2017context} proposed a dynamic mechanism to regulate the ratios at which source and target contexts contribute to the generation of target words via context gates. 
\citet{fu2023decoder} identified the attention degeneration problem, i.e., as the generation step number grows, less and less attention is focused on the source sequence, in language models.
Numerous methods have been proposed to enhance faithfulness in natural language generation~\citep{li2022faithfulness}, all of which were designed to address unfaithfulness under the encoder-decoder architecture neural network.
With the emergence of decoder-only LLMs, our focus shifted to the coverage of LLM-based translation. Although these approaches were originally designed for encoder-decoder models and cannot be directly applied to LLMs, they provide a valuable guide for current work on decoder-only language models~\citep{chen2023improving,he2024exploring}.
The research most similar to ours is that of \citet{chen2023improving}. They also focus on the unfaithful translation of LLMs, but their methods of addressing these problems differ significantly from ours. They enhance instruction comprehension by adding an instruction representation to the subsequent input and response representations, which are tailored to the tuning application scenario. In contrast, our methods fit different application scenarios. Reweighting attention works in low-resource settings without a full parallel corpus. Contrastive decoding fits API-only LLMs without internal information access. Target-instruction tuning suits tuning scenarios without adding extra modules or parameters.

\subsection{LLMs for machine translation}
Recently, several studies focused on how to prompt LLMs for machine translation~\citep{zhang2023prompting,vilar-etal-2023-prompting,agrawal-etal-2023-context}. \citet{zhu2023multilingual} evaluated the performance of eight prominent large language models, including ChatGPT and GPT-4~\citep{openai2023gpt4}, in multilingual machine translation. \citet{jiao2023ischatgpt} conducted a preliminary evaluation of ChatGPT's machine translation abilities, exploring translation prompts, multilingual translation, and robustness. ChatGPT competes well with commercial products in high-resource European languages but faces challenges with low-resource or distant languages. Parrot~\citep{jiao-etal-2023-parrot}, a framework for enhancing chat-based translation, utilizes open-source LLMs, human-written translations, and feedback data to address challenges posed by restricted APIs. \citet{xu2023paradigm} proposed a novel fine-tuning approach for LLMs that is specifically designed for the translation task and achieves significant advancement on LLM-based machine translation compared to the previous attempts. 
While LLMs have showcased remarkable performance, they inevitably confront various challenges in practical applications, with hallucinations emerging as one of the most notable issues~\citep{wang2023survey,zheng2023does,zhang2023sirens}. \citet{10.1162/tacl_a_00615} examine hallucinations in large multilingual translation models, conducting a comprehensive analysis on both conventional M2M neural machine translation models~\citep{JMLR:v22:20-1307} and ChatGPT, shedding light on the unfaithfulness translation brought by LLMs. 

However, previous studies have not explored the relationship between unfaithful or hallucinatory translation and the contribution from input tokens to generated tokens in LLMs. We have attempted to fill this gap. Our analysis of the contribution scores of input tokens has allowed us to highlight the limitations inherent in the decoder-only architecture of LLMs, specifically the lack of an explicit cross-attention module. Such a restriction can lead to inadequate focus on the source context, which we believe might potentially increase the risk of unfaithfulness in LLMs' generations. 

\section{Conclusion}
In this research, we identify the issue of insufficient focus on source context in LLMs when applied to machine translation tasks and accordingly, we propose the reweight attention to adjust the attention weight of source context to help models focus on the source context during generation, contrastive decoding to reduce the influence of target prefixes, and target-constrained tuning to encourage LLMs to avoid excessive dependence on specific target prefixes. Our experimental results show marked improvements in translation performance across several language pairs in our proposed unfaithful translation test sets, outperforming baseline methods and effectively reducing the phenomenon of hallucinatory and unfaithful translations. Both our quantitative and qualitative analysis of contribution scores indicate the significance of our proposed methods in addressing the identified issue.  
While we only explore related issues in the application of machine translation, it is natural to extend our methods to other seq2seq tasks(e.g., summarization), which we leave for future exploration.

\section*{Acknowledgment}
We sincerely thank all the reviewers for their valuable and insightful comments, which have greatly enhanced the quality of our work. The work was partial supported by the National Natural Science Foundation of China under Grant No. 62276077, Guangdong Basic and Applied Basic Research Foundation under Grant No. 2024A1515011205, Shenzhen College Stability Support Plan under Grants XWD20220811170358002 and GXWD20220817123150002. The work of Yang Xiang was supported by the National Natural Science Foundation of China under Grant No. 62106115 and Major Key Project of PCL under Grant No. PCL2022D01.

\section*{Limitations}
This research conducts a preliminary investigation into the hallucinatory and unfaithful translations resulting from insufficient focus on the source context in decoder-only LLMs, and the following aspects can be improved upon in future work:
\begin{itemize}
    \item \textbf{Variations in Instructions:} In our study, we did not consider the effects of varying the instruction prompt, nor did we examine the impact of our proposed methods under different instructions.
    \item \textbf{Testing Limited to Greedy Search:} Despite the availability of numerous decoding strategies and generation configurations, we set the temperature to 0 and 'do\_sample' to False to demonstrate our proposed methods' effectiveness. We did not investigate the performance of these methods in combination with other generation strategies, such as beam search, top-k sampling, or nucleus sampling.
    \item \textbf{Increased Latency} Although our proposed methods are effective, they incur a higher computational cost compared to the standard settings. The contribution-based alignment selection strategy requires extra time to compute the dot product. Target-constrained tuning necessitates two forward passes, nearly doubling the training time.
\end{itemize}

\section*{Ethical Considerations}
In our human-collected datasets, translations are obtained using zero-shot prompting from open-sourced LLMs, and thus any problematic responses can be attributed to the organizations that release these LLMs. We do not anticipate any significant risks associated with our research. In theory, our framework for mitigating unfaithful or hallucinatory translation could yield higher-quality translations without any toxic content. Based on our observations, our proposed methods have not resulted in any detrimental responses.
To ensure the reproducibility of our experiments, we intend to make our code and evaluation data available to the public.


\bibliography{acl_latex}
\clearpage

\appendix
\section{Analyzing Contribution of input tokens}
\label{alti_sec}
In each layer, the attention block computations can be expressed simply as a linear function of the input representations. Given a model with $H$ heads, the attention block output of the $i$-th token $y_i$ is computed by applying the layer normalization (LN) over the sum of the residual vector $x_i$ and the output of the multi-head attention module (MHA) $\hat{x_i}$.
\begin{equation}
y_i = \text{LN}(\hat{x_i} + x_i)
\label{ln_res}
\end{equation}
After the MHA module, $\hat{x_i}$ can be expressed as the linear combination of different input tokens and different attention heads:
\begin{equation}
\hat{x}_i = \sum_h^H W_o^h \sum_j^J A_{i,j}^h W_v^h x_j + b_o
\end{equation}
Given a vector $u$, $\text{LN}(u)$ can be reformulated as $\frac{1}{\sigma(u)}Lu + \beta$, where $L$ is a linear transformation. 
By swapping summations and utilizing the linearity of LN, we can now rewrite Eq. \ref{ln_res} as:
\begin{equation}
y_i = \sum_j T_i(x_j) + \frac{1}{\sigma(\hat{x_i} + x_i)} L b_o + \beta
\end{equation}
where the transformed vectors \(T_i(x_j)\) are:
\begin{equation}
\resizebox{\linewidth}{!}{
$T_i(x_j) = \begin{cases} \frac{1}{\sigma(x_i+x_j)}L\left(\sum_h W_o^h A_{ij}^h W_v^h x_j\right) & \text{if } i \neq j \\\ \frac{1}{\sigma(\hat{x_i}+x_i)}L\left(\sum_h W_o^h A_{ii}^h W_v^h x_j + x_i\right)  & \text{if } i = j \end{cases}$
}
\end{equation}
\citet{kobayashi-etal-2020-attention,kobayashi-etal-2021-incorporating} propose assessing the contribution of each input vector $x_j$ to the layer output $y_i$ through the Euclidean norm: $c_{i,j} = \|T_i(x_j)\|_2$. While \citet{ferrando-etal-2022-towards} argue that transformed representations exhibit reduced anisotropy and they suggest using $l_1$ norm: $c_{i,j} = \| y_i - T_i(x_j) \|_1 $. Normalizing these contributions yields a layer-wise contribution matrix $C \in \mathbb{R}^{J \times J}$. By employing a similar method as attention rollout~\citep{abnar-zuidema-2020-quantifying}, an overall contribution matrix for input tokens to the generated token $y_i$ is obtained.

While previous studies have investigated the contributions of input tokens in machine translation~\citep{ferrando-etal-2022-measuring}, their focus was exclusively on encoder-decoder style transformers. 
To the best of our knowledge, we are the first to analyze the contribution of input tokens in LLMs. Taking into account both the orientation and norm of the transformed vectors, we modify $c_{i,j}=\frac{T_i(x_j) \cdot y_i}{||y_i||_2} $ to represent the contribution of transformed vectors towards the generated tokens. A larger vector projection onto $y_i$ is expected to indicate a higher contribution. 
Given that future tokens are masked in the Transformer decoder, there is an inherent bias toward the initial tokens of the input sequence. Direct aggregation of each layer-wise contribution matrix may further intensify this bias~\citep{abnar-zuidema-2020-quantifying}. Therefore, to prevent this, we normalize each layer-wise contribution matrix before aggregation.

\section{Experimental Setting}
\label{appendix_exp_setting}
\subsection{Instruction prompts}
\paragraph{Translation prompt for LLaMA2-chat model.}
Our translation approach builds upon the work of \citet{sennrich2023mitigating}
The input to Llama2-chat consists of a \textcolor{brown}{system prompt} and an \textcolor{blue}{instruction}. To ensure that the assistant's response begins with the actual translation rather than an introductory phrase or prologue, we force-decode the \textcolor{orange}{prefix} of the assistant response.
Here is the zeroshot translation instruction prompt:
\\
<s>[INST] <<SYS>>\\
\textcolor{brown}{You are a machine translation system that translates sentences from English to German. You just respond with the translation, without any additional comments.}\\
<</SYS>>\\
\textcolor{blue}{Sie stehen keine 100 Meter voneinander entfernt: Am Dienstag ist in Gutach die neue B 33-Fußgängerampel am Dorfparkplatz in Betrieb genommen worden - in Sichtweite der älteren Rathausampel.\\
Translate to English} [/INST]\textcolor{orange}{Sure, here’s the translation:}

In the case of fewshot translation prompt, we adopt the same fewshot strategy used in \citet{zhu2023multilingual}. We use eight randomly sampled translation pairs from the respective training set as in-context exemplars. These exemplars are presented in “<X>=<Y>” format, where “<X>” and “<Y>” are the placeholder for the source and target sentence. Line-break serves as the exemplar's concatenation symbol.

Similar to traditional translation systems, we use bilingual sentence pairs to instill basic translation capabilities into LLMs. We adopt the Stanford Alpaca method~\citep{alpaca} to convert bilingual sentence pairs into an instruction-following format, which fine-tunes LLMs for translation tasks. 
For the instruction tuning prompt of the LLaMA2-chat model, we retained the zeroshot translation instruction prompt, but we did not compute the loss for the \textcolor{brandeisblue}{instruction query} part. Instead, we only calculated the loss for the \textcolor{red}{response output label}, as illustrated in the subsequent examples of the translation instruction prompt:
\\
<s>[INST] <<SYS>>\\
\textcolor{brandeisblue}{
You are a machine translation system that translates sentences from English to German. You just respond with the translation, without any additional comments.}\\
<</SYS>>\\
\textcolor{brandeisblue}{Zwei Anlagen so nah beieinander: Absicht oder Schildbürgerstreich?\\
Translate to English} [/INST]\textcolor{brandeisblue}{Sure, here’s the translation:}\textcolor{red}{Two sets of lights so close to one another: intentional or just a silly error?}\\

\subsection{Evaluation Data}
\paragraph{Unfaithful Translation Data.}
Specifically, we use our contribution scores analysis tool adapted for LLMs, which is modified from the ALTI+\footnote{https://github.com/mt-upc/transformer-contributions-nmt} method, to filter the data. If the source text contributions minus the target prefixes' contributions fall below a certain threshold, we collect them. We apply our methods to filter evaluation data on publicly available parallel data, such as News-Commentary v16 for German to English (De$\Leftrightarrow$En) and TED2013 for Chinese to English (Zh$\Leftrightarrow$En)\footnote{https://opus.nlpl.eu}. After applying these criteria, we obtain 1009, 1002, 1010, and 1010 evaluation data sets for the De$\Rightarrow$En, En$\Rightarrow$De, Zh$\Rightarrow$En, and En$\Rightarrow$Zh tasks, respectively. Given our choice of translation instances with low source token contributions, which results in a dataset that contains instances that either deviate from the original sentence or lack semantic connection (e.g., copied instructions, text continuation, hallucinatory translation). 
\paragraph{General Data.}We evaluate the translation performance of LLMs on two sources of test sets:
\begin{itemize}
    \item \textbf{Flores-101:} We use the Flores-101 which serves as the evaluation benchmark for multilingual-machine systems and the number of test samples is 1012 for all translation directions~\citep{flores101}.
    \item \textbf{WMT22 Test sets:} We also utilize the test sets from the WMT22 competition. These sets are constructed based on recent content from various domains, including news, social, e-commerce, and conversational domains. The sample numbers for the De$\Rightarrow$En, En$\Rightarrow$De, Zh$\Rightarrow$En, and En$\Rightarrow$Zh tasks are 1984, 2037, 1875, and 2037, respectively~\citep{kocmi-etal-2022-findings}.
\end{itemize}
\subsection{Model Training}
We conduct our main experiments with HuggingFace Transformers\footnote{https://github.com/huggingface/transformers} on open-source LLMs from the LLaMA2 family~\citep{touvron2023llama}. Specifically, we choose LLaMA2-7b-chat\footnote{https://huggingface.co/meta-llama/Llama-2-7b-chat-hf}, BLOOMZ-7b1-mt\footnote{https://huggingface.co/bigscience/bloomz-7b1-mt}~\citep{muennighoff-etal-2023-crosslingual}, ChatGLM3-6b\footnote{https://huggingface.co/THUDM/chatglm3-6b}~\citep{du-etal-2022-glm}, and Vicuna-7b\footnote{https://huggingface.co/liuhaotian/llava-v1.6-vicuna-7b}~\citep{vicuna2023} with matched parameters, and also include LLaMA2-7b-chat, LLaMA2-13b-chat and LLaMA2-70b-chat to study the effect of model sizes. 
The hyperparameters used for finetuning are mainly aligned with those of Stanford Alpaca\footnote{https://github.com/tatsu-lab/stanford\_alpaca}~\citep{alpaca}. 
For instruction tuning and target-constrained tuning, we finetune models over 5 epochs with a learning rate of 1e-4, using the corresponding language direction parallel data. During the implementation of the LoRA finetuing\footnote{https://github.com/tloen/alpaca-lora}, we set the `lora\_r' to 16, `lora\_alpha' to 32, `lora\_dropout' to 0.3. The `target\_modules' are configured to include the query, key, value projection and output projection within the attention module. For more specific hyperparameters, please refer to our released scripts. We perform the finetuning process on 8 Nvidia A100 GPUs and employ DeepSpeed\footnote{https://github.com/microsoft/DeepSpeed} ZeRO stage 2 for model parallel.

\section{Ablation Study}
\label{appendix_ablation}
We analyze specific factors related to our proposed methods that may impact the translation performance of LLMs. As a default setting, we perform ablation studies on our proposed unfaithful translation dataset using the LLaMA2-7b-chat model.
\subsection{Reweight Attention different strategy and the effect of the scale factor $\omega$ and window size}
\begin{table}[!ht]
\centering
\begin{tabular}{ccc}
\hline
strategy               & BLEU & COMET \\ \hline
baseline zeroshot      & 11.4 & 74.3  \\
Monotonic Local Window & 11.8 & 74.7  \\
Heuristic Local Window & 12.5 & 75.0  \\
Global Window          & 12.4 & 75.0  \\ \hline
\end{tabular}
\caption{different strategies for the reweight attention}
\label{reweight_strategies}
\end{table}
\begin{figure}
    \centering
    \includegraphics[width=1\linewidth]{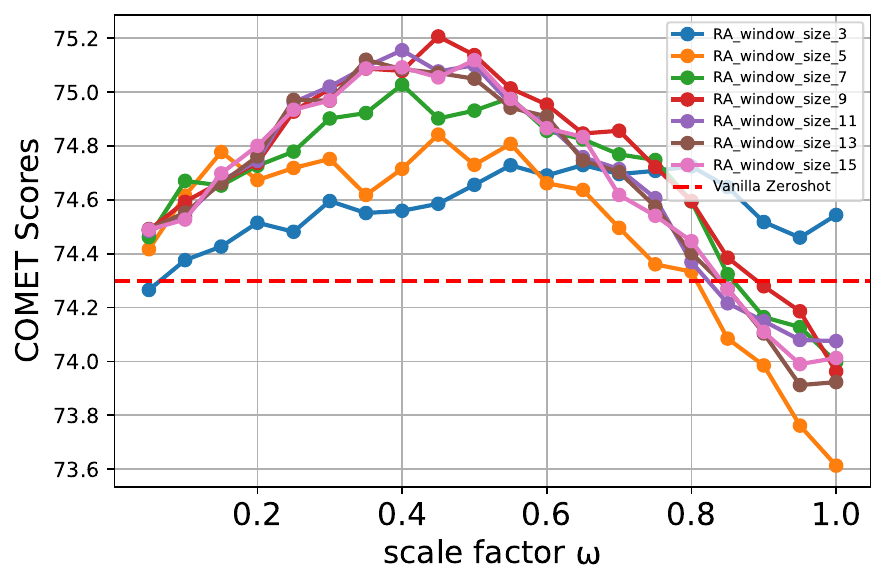}
    \caption{COMET scores over scale factors $\omega$ and window size $D$ parameters of reweight attention}
    \label{ra_ablation}
\end{figure}
Two strategies for choosing local window anchors to adjust attention scores. For comparison, we have also introduced a global window strategy that adjusts the attention scores across the entire source context. 
In this study, we aim to determine which strategy most effectively improves the translation performance of LLMs. 
As shown in Table \ref{reweight_strategies}, our heuristic strategy, which prioritizes the most significant source tokens based on their contribution, outperforms the other strategies. The monotonic strategy is the most subtle optimization strategy, as aligning the source and target tokens in a step-by-step manner may not conform to a human-like translation. 
Interestingly, adjusting the attention scores across the entire window of source tokens produces results comparable to the heuristic approach. We further investigated the translation performance across varying local window sizes and scale factor $\omega$, as shown in Figure \ref{ra_ablation}. Our results suggest that significant performance enhancement is observed with larger windows when the window size is initially small(e.g. 3 or 5). However, as the window size increases beyond a certain point(e.g. 9), we notice a slight decrease in translation performance. 
This can be attributed to the fact that incorporating irrelevant source information, not related to the currently generated token, may introduce additional noise.
Additionally, assigning a small factor does not effectively shift the model's focus toward the source tokens. Conversely, overemphasis on the source context can lead to performance decline. In total, the optimal balance appears to be a window size of around 9 and a scale factor of around 0.5.
\subsection{Effect of adjustment level $\alpha$ in Contrastive Decoding}
\begin{figure}
    \centering
    \includegraphics[width=1\linewidth]{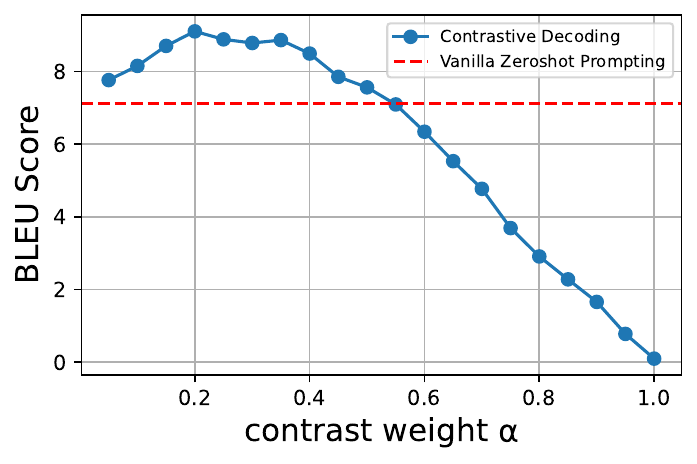}
    \caption{BLEU scores over the various penalty degrees $\alpha$}
    \label{cd_ablation}
\end{figure}
In our proposed contrastive decoding method, we introduce a hyperparameter $\alpha$ to control the penalty degree of contrastive decoding. A smaller $\alpha$ results in the model predictions' output distribution being closer to the original distribution of the next tokens. We carry out experiments using various values of $\alpha$ and present the results in Figure \ref{cd_ablation}. Smaller values of $\alpha$(e.g., 0.1) do not yield performance as robust as larger values $\alpha$(e.g., 0.5), suggesting that the models continue to generate unconditioned tokens when a penalty of a lower degree is utilized. However, increasing the value of $\alpha$(e.g., 1) further causes a decline, due to the unfluent and ungrammatical generation from erasing too much target contribution.
\subsection{Effect of mask ratio $\beta$ and KL-divergence coefficient $\lambda$ in target-constrained tuning}
\begin{figure}
    \centering
    \includegraphics[width=1\linewidth]{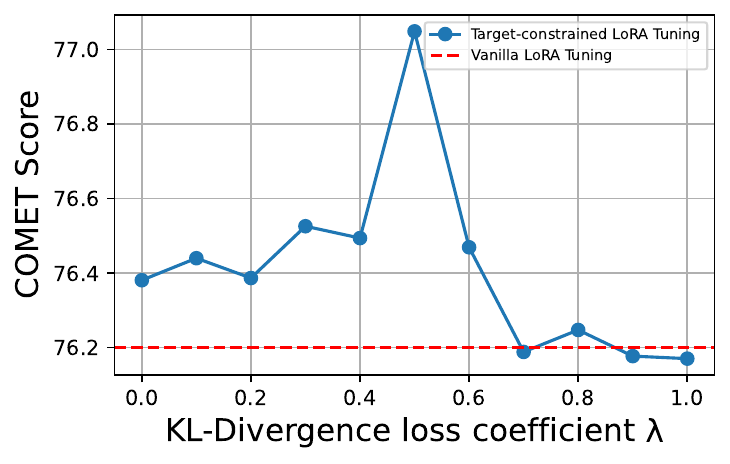}
    \caption{COMET scores over different the KL Divergence coefficients $\lambda$}
    \label{kl_div_ablation}
\end{figure}
\begin{figure}
    \centering
    \includegraphics[width=1\linewidth]{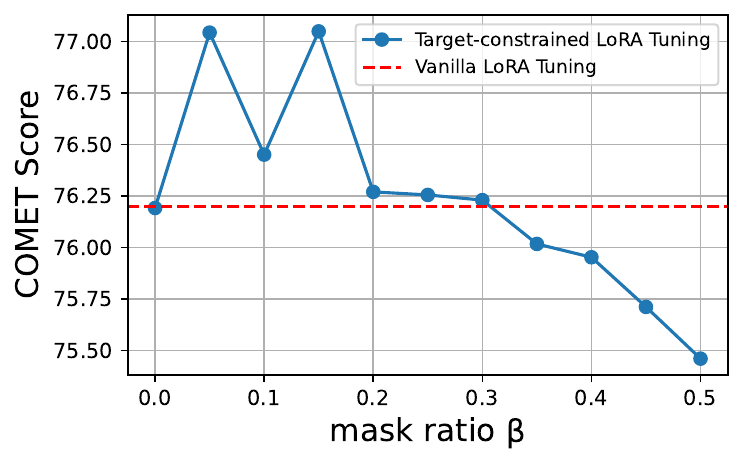}
    \caption{COMET scores over different mask ratios $\beta$}
    \label{mask_ratio_ablation}
\end{figure}
We study the influence of two parameters: the mask ratio $\beta$ and the weight assigned to the KL-divergence loss $\lambda$.  
Here, we let $\beta$ vary between $\{0 \rightarrow 0.5\}$ and $\lambda$ between $\{0 \rightarrow 1\}$ and 
From the results in Figure \ref{kl_div_ablation} and \ref{mask_ratio_ablation}, we find that: 1) A mask ratio of 0.15 and a KL coefficient of 0.5 yields the best performance; 2) Target-constrained tuning consistently achieves strong results with a mask ratio 0.05 and 0.15; 3) As the mask ratio increases, target-constrained tuning fails to converge because the model struggles with the high randomness from the masked target prefix; 4) Smaller $\lambda$ values underperform compared to larger ones, highlighting the need for attention to KL-divergence regularization. However, over-regularization is detrimental. The best balance is achieved at $\lambda=0.5$. 

\section{Human Evaluation}
\begin{table}[]
\centering
\begin{tabular}{ccc}
\hline
\multicolumn{3}{c}{\textbf{Fleiss’s Kappa Scores}}                            \\
Correct & Hallucination & Errors \\
\hline
0.95      & 0.79           & 0.88                 \\
\hline
\end{tabular}
\caption{Fleiss’s Kappa inter-annotator agreement scores for the three annotation categories.}
\label{inter-annotator}
\end{table}
In this appendix, we describe the manual evaluation. First, we detail the simple guidelines that were presented to manual annotators. Second, we report the number of annotators and inter-annotation agreements.
\label{appendix_human_eval}
\paragraph{annotation guidelines}
\begin{table*}[]
\centering
\begin{tabular}{lp{12cm}}
\hline
Annotation Types &
  Definition \\ \hline
Correct &
  The translation fully conveys the meaning of the original text. It may contain content that does not affect the availability of the content or Minor understandability errors (eg: incorrect punctuation conversion) \\
Omission &
  Translation is an example of omission if and only if part of the source language sentence is translated correctly but the remaining part is not (not including an attempt to translate that part but the translation is wrong, which is a complete lack of attempt) to translate that part of the content) \\
Semantically Detached &
  Some of these incorrect translations are supported by the content of the source language sentences. However, a large proportion of this mistranslation is not (it conveys a different meaning than the one in the source sentence). \\
Off-target &
  An example of off-target translation is when the translation system fails to translate the source language into the target language, that is, non-target language fragments appear in the translation. \\
Named-Entity &
  Named-entities mistranslation are mistranslated (for example: mistranslation of names of people, places, organizations, dates, prices, etc.) \\
Over-translation &
  A translation is an example of over-translation if and only if all the contents of the source language sentence are translated correctly, but the translation system excessively generates more translations. \\ \hline
\end{tabular}
\caption{Human annotations Guidelines}
\label{guidelines}
\end{table*}
The annotators were provided with the guidelines as outlined in Table \ref{guidelines}. 
We consolidate the labels based on a majority vote. Notably, we also designate unfaithful translations, encompassing under-translations and over-translations, as hallucinations. Translations that exhibit semantic deviation from the source context are also recognized as hallucinations.
Concretely, for the purpose of reporting, we grouped `Named-entity mistranslation' and `Off-target' under the `Error' category, while `Semantically detached', `Omission', and `Over-translation' were classified under the `Hallucination' category."

\paragraph{Inter-annotation agreement}
To ensure the reliability and quality of our annotated unfaithful translation datasets, we additionally assigned the same set of another 100 randomly sampled translation instances to two annotators.
The Fleiss' Kappa statistic, representing the agreement in the assessment of annotation categories between the annotators, is presented in Table \ref{inter-annotator}. 
As demonstrated, we can effectively classify the different types of translation with strong agreement between the two annotators, thereby indicating the effectiveness of the human evaluation test set. 
This confirms the suitability of our annotated data for our analysis.
\begin{table*}[!ht]
\centering
\resizebox{\linewidth}{!}{
\begin{tabular}{cccccc}
\hline
Method & \begin{tabular}[c]{@{}c@{}}semantically\\ detached\end{tabular} & omission & over-translation & off-target & name-entity \\ \hline
Vanilla zeroshot           & 28 & 18 & 18 & 9 & 6 \\
Reweight attention         & 11 & 8  & 6  & 8 & 6 \\
Contrastive decoding       & 12 & 11 & 3  & 8 & 6 \\
Vanilla instruction tuning & 15 & 3  & 3  & 3 & 6 \\
Target-constrained tuning  & 8  & 1  & 0  & 3 & 5 \\ \hline
\end{tabular}}
\caption{The detailed categories of the collected 100 examples in Section\ref{human_eval}.}
\label{detailed_categories_human_eval}
\end{table*}
\begin{table*}[!ht]
\centering
\resizebox{\linewidth}{!}{
\begin{tabular}{ccccc}
\hline
Method & BLEU & COMET & BLEURT & Human Evaluation Numerical Score \\ \hline
Vanilla zeroshot           & 7.45 & 65.5 & 49.3 & -14.2 \\
Reweight attention         & 11.3 & 68.7 & 53.1 & 43.65 \\
Contrastive decoding       & 11.7 & 69.2 & 52.6 & 40    \\
Vanilla instruction tuning & 14.8 & 74.7 & 58.9 & 57.85 \\
Target-constrained tuning  & 15.4 & 75.0 & 60.5 & 70.5  \\ \hline
\end{tabular}
}
\caption{The numerical scores of human evaluation and automatic metrics on the sampled set}
\label{numerical_scores_consistency}
\end{table*}
\section{Connection between Human Evaluation and Automatic metrics}
\label{connection_human_eval_automatic_metrics}
To reveal the consistency of hallucination evaluation between automated metrics and human evaluation in cases of unfaithful translation, we conduct further analysis. This analysis focuses on the correlation between human evaluations quantified by numerical scores and automated metrics.
According to Table~\ref{guidelines} in Appendix~\ref{appendix_human_eval}, translations are first categorized into one of six annotation categories, and the detailed categories of the sampled set are shown in Table~\ref{detailed_categories_human_eval}. To convert these categories into numerical scores, we can use the proportion of hallucinatory translations to measure hallucination severity, following~\citet{dale-etal-2023-detecting}. A higher proportion indicates more severe hallucinations. Alternatively, to emphasize unfaithfulness in translation, we assign a score of -0.4 for omission and over-translation, a score of -0.25 for semantically detached, a score of -0.2 for error translations, and a score of +1 for correct translations.

We first convert human evaluation categories into numerical scores using the previously discussed method, and then compute BLEU, COMET, and  BLEURT~\citep{sellam2020bleurt} metrics for these examples. The results are presented in the Table~\ref{numerical_scores_consistency}

Finally, we calculated the Pearson correlation coefficient between the automated metrics and human evaluation scores. The Table~\ref{pearson_correlation} shows that the BLEU metric aligns most closely with human evaluation scores, followed by COMET. Thus, we further confirm that BLEU and COMET are proper for our experiments as they can reflect the level of unfaithful translation to a certain degree.
\begin{table}[]
\centering
\resizebox{\linewidth}{!}{
\begin{tabular}{cc}
\hline
Automatic\_metric & Pearson Correlation Coefficient \\ \hline
BLEU              & 0.9640                          \\
COMET             & 0.8942                          \\
BLEURT            & 0.8907                          \\ \hline
\end{tabular}
}
\caption{The Pearson correlation coefficient between automatic metrics and human evaluation numerical score.}
\label{pearson_correlation}
\end{table}

\section{Analysis of the imbalanced improvement phenomenon}
\label{analysis_imbalanced}
\begin{table}[]
\centering
\resizebox{\linewidth}{!}{
\begin{tabular}{cccc}
\hline
Contribution Metric                       & Unfaithful dataset & WMT22  & Flores101 \\ \hline
Llama2-7b-chat average contribution score & 0.7914             & 0.8381 & 0.8434    \\ \hline
\end{tabular}
}
\caption{The contribution metric scores between different test sets.}
\label{contribution_scores_imbalanced}
\end{table}
Our paper focuses on unfaithful translations in LLMs caused by inadequate attention to the source context. Therefore, we collect data containing such issues to form a specific dataset and conduct main experiments on this dataset, with results in Table~\ref{main_result} showing our methods effectively address the unfaithful translation issue. Only then did we generalize our methods to open general datasets, with results in Table 2. Experimental results show improvement of our method is less significant on general datasets than on the unfaithful translation test set. This is mainly due to the rare unfaithful translations when using the Llama-2-7b-chat model for zero-shot translation on the Flores101 and WMT22 test sets. We also calculate source contribution scores, as detailed in Appendix A (higher scores mean more focus on the source context during LLM generation, thus is less possible to generate unfaithful translations), across both general and proposed test sets to illustrate this point. As shown in the Table~\ref{contribution_scores_imbalanced}, the source contribution scores on the general dataset are higher than the proposed dataset, which explains why our method's improvements are less significant on general datasets.

\section{Further analysis of the sudden drop}
\label{analysis_sudden_drop}
\begin{table}[]
\centering
\resizebox{\linewidth}{!}{
\begin{tabular}{cccccc}
\hline
mean & min & max & median & mode & std   \\ \hline
39.9 & 7   & 300 & 33     & 22   & 26.94 \\ \hline
\end{tabular}
}
\caption{The basic statistics of the source length in the Zh-to-En test set.}
\label{len_stats}
\end{table}
The lines in the figure represent averages not for all examples but for those corresponding to examples with at least $i$ tokens. Therefore, when the number of tokens exceeds 100, only translations with a sentence length greater than 100 tokens are taken into account.
The basic statistics of the source length in the Zh-to-En test set are presented in Table~\ref{len_stats}.

We conducted the human evaluation to the long source context cases, and found that when the source context length exceeds 200, translations produced by LLM using vanilla zero-shot settings show more instances of not following instructions and more omissions, resulting in reduced reliance on the source context during translation generation. Consequently, the contribution of the source context diminishes.

\section{Extended comparison experiments between model families and scales within WMT22 test set and unfaithful test set}
\label{more_comparison_exps}

In the comparison of these language model families, ChatGLM3-6b demonstrates superior performance in the English to Chinese (En $\Rightarrow$ Zh) language translation direction, largely as a result of its enhanced modeling of Chinese during the pretraining phase. 
On the other hand, BLOOMZ-7b1-mt performs better in the Chinese to English (Zh $\Rightarrow$ En) language translation direction. 
This enhancement can be ascribed to its substantial exposure to a varied compilation of parallel multilingual pretraining corpora, coupled with the prevalence of English in the pretraining corpus of BLOOM. 

As evidenced by the results shown in Table~\ref{model_size_unfaithful}, the 7b model experiences the most significant boost, while larger models only achieve a marginal enhancement. Given that the test unfaithful translation sets are collected from the poorest translation instances of the 7b chat model, larger models demonstrate superior translation capabilities and generate fewer unfaithful issues within such a dataset. This inconsistent improvement is primarily due to the difference in the number of unfaithful issues generated by LLMs of varying scales. 

\begin{table*}[!ht]
\centering
\resizebox{\linewidth}{!}{\begin{tabular}{ccccccccc}
\hline
\multirow{2}{*}{System} &
  \multicolumn{2}{c}{En $\Rightarrow$ De} &
  \multicolumn{2}{c}{De $\Rightarrow$ En} &
  \multicolumn{2}{c}{En $\Rightarrow$ Zh} &
  \multicolumn{2}{c}{Zh $\Rightarrow$ En} \\ \cline{2-9} 
                                & BLEU & COMET & BLEU & COMET & BLEU & COMET & BLEU & COMET \\ \hline
\multicolumn{9}{c}{Base Model:BLOOMZ-7b1-mt}                                                \\
Vanilla Instruction tuning LoRA & 20.5 & 64.2  & 31.8 & 76.6  & 44.6 & 84.4  & 25.5 & 79.1  \\
Target-constrained tuning LoRA  & 21.2 & 64.4  & 32.5 & 76.7  & 45.2 & 84.6  & 26.2 & 79.4  \\ \hline
\multicolumn{9}{c}{Base Model:ChatGLM3-6b}                                                  \\
Vanilla Instruction tuning LoRA & 27.1 & 73.2  & 37.6 & 81.0  & 46.4 & 84.4  & 26.8 & 79.1  \\
Target-constrained tuning LoRA  & 27.2 & 73.2  & 37.8 & 81.1  & 46.7 & 84.5  & 27.2 & 79.2  \\ \hline
\multicolumn{9}{c}{Base Model:Vicuna-7b}                                                    \\
Vanilla Instruction tuning LoRA & 32.5 & 80.6  & 39.8 & 82.0  & 42.0 & 82.6  & 25.0 & 76.8  \\
Target-constrained tuning LoRA  & 33.5 & 81.3  & 41.4 & 83.1  & 42.9 & 83.0  & 25.2 & 76.9  \\ \hline
\end{tabular}}
\caption{Translation performance of various families of
LLMs with a similar size on WMT22 test sets.}
\label{model_families_wmt22_full}
\end{table*}

\begin{table*}[!ht]
\centering
\resizebox{\linewidth}{!}{\begin{tabular}{ccccccccc}
\hline
\multirow{2}{*}{System} &
  \multicolumn{2}{c}{En $\Rightarrow$ De} &
  \multicolumn{2}{c}{De $\Rightarrow$ En} &
  \multicolumn{2}{c}{En $\Rightarrow$ Zh} &
  \multicolumn{2}{c}{Zh $\Rightarrow$ En} \\ \cline{2-9} 
                                & BLEU & COMET & BLEU & COMET & BLEU & COMET & BLEU & COMET \\ \hline
\multicolumn{9}{c}{Base Model:LLaMA2-7b-chat}                                               \\
Vanilla Instruction tuning LoRA & 34.0 & 81.7  & 42.0 & 83.2  & 38.9 & 81.6  & 25.5 & 78.3  \\
Target-constrained tuning LoRA  & 34.5 & 81.8  & 42.5 & 83.4  & 39.2 & 81.8  & 25.8 & 78.4  \\ \hline
\multicolumn{9}{c}{Base Model:LLaMA2-13b-chat}                                              \\
Vanilla Instruction tuning LoRA & 37.0 & 83.2  & 43.0 & 83.5  & 43.7 & 84.2  & 27.6 & 79.3  \\
Target-constrained tuning LoRA  & 37.6 & 83.4  & 43.8 & 83.9  & 43.9 & 84.2  & 27.8 & 79.4  \\ \hline
\multicolumn{9}{c}{Base Model:LLaMA2-70b-chat}                                              \\
Vanilla Instruction tuning LoRA & 41.3 & 84.8  & 46.0 & 84.7  & 49.2  &  85.6  & 30.2 & 80.5  \\
Target-constrained tuning LoRA  & 41.8 & 85.0  & 46.9 & 84.9  &        49.9  & 86.0 & 31.1  & 80.8      \\ \hline
\end{tabular}}
\caption{Translation performance of various families of
LLMs with a similar size on WMT22 test sets.}
\label{model_size_wmt22_full}
\end{table*}

\begin{table*}[!ht]
\centering
\resizebox{\linewidth}{!}{\begin{tabular}{ccccccccc}
\hline
\multirow{2}{*}{System} &
  \multicolumn{2}{c}{En $\Rightarrow$ De} &
  \multicolumn{2}{c}{De $\Rightarrow$ En} &
  \multicolumn{2}{c}{En $\Rightarrow$ Zh} &
  \multicolumn{2}{c}{Zh $\Rightarrow$ En} \\ \cline{2-9} 
                                & BLEU & COMET & BLEU & COMET & BLEU & COMET & BLEU & COMET \\ \hline
\multicolumn{9}{c}{Base Model:BLOOMZ-7b1-mt}                                                \\
Vanilla Instruction tuning LoRA & 11.9 & 62.7  & 21.1 & 75.3  & 20.9 & 78.3  & 16.2 & 77.5  \\
Target-constrained tuning LoRA  & 12.0 & 62.7  & 21.9 & 76.0  & 21.2 & 78.4  & 17.1 & 77.8  \\ \hline
\multicolumn{9}{c}{Base Model:ChatGLM3-6b}                                                  \\
Vanilla Instruction tuning LoRA & 15.4  & 70.2   & 25.8  & 80.4   & 20.5 & 78.5  & 16.3 & 77.2  \\
Target-constrained tuning LoRA  & 15.9  & 70.3   & 26.2  & 80.5   & 21.3 & 78.9  & 17.0 & 77.5  \\ \hline
\multicolumn{9}{c}{Base Model:Vicuna-7b}                                                    \\
Vanilla Instruction tuning LoRA & 15.6 & 69.9  & 24.6 & 78.7  & 20.5 & 77.7  & 14.2 & 71.6  \\
Target-constrained tuning LoRA  & 16.0 & 70.1  & 25.8 & 79.5  & 21.1 & 77.9  & 14.8 & 71.7  \\ \hline
\end{tabular}}
\caption{Translation performance of various families of
LLMs with a similar size on human-collected unfaithful translation test sets.}
\label{model_families_unfaithful}
\end{table*}

\begin{table*}[!ht]
\centering
\resizebox{\linewidth}{!}{\begin{tabular}{ccccccccc}
\hline
\multirow{2}{*}{System} &
  \multicolumn{2}{c}{En $\Rightarrow$ De} &
  \multicolumn{2}{c}{De $\Rightarrow$ En} &
  \multicolumn{2}{c}{En $\Rightarrow$ Zh} &
  \multicolumn{2}{c}{Zh $\Rightarrow$ En} \\ \cline{2-9} 
                                & BLEU & COMET & BLEU & COMET & BLEU & COMET & BLEU & COMET \\ \hline
\multicolumn{9}{c}{Base Model:LLaMA2-7b-chat}                                               \\
Vanilla Instruction tuning LoRA & 18.6 & 77.4  & 28.7 & 79.8  & 18.6 & 76.2  & 15.5 & 76.5  \\
Target-constrained tuning LoRA  & 20.0 & 77.9  & 30.1 & 81.1  & 19.1 & 76.5  & 16.6 & 77.0  \\ \hline
\multicolumn{9}{c}{Base Model:LLaMA2-13b-chat}                                              \\
Vanilla Instruction tuning LoRA & 21.3 & 80.4  & 30.7 & 81.5  & 20.0 & 77.8  & 17.7 & 77.9  \\
Target-constrained tuning LoRA  & 21.9 & 80.6  & 31.3 & 81.7  & 20.6 & 78.0  & 18.2 & 78.1  \\ \hline
\multicolumn{9}{c}{Base Model:LLaMA2-70b-chat}                                              \\
Vanilla Instruction tuning LoRA & 23.1 & 81.4  & 33.4 & 84.5  & 	21.7 & 79.1   & 19.7  & 79.0   \\
Target-constrained tuning LoRA  & 23.8 & 81.8  & 33.9 & 84.8  & 22.1  & 79.3   & 20.4 & 79.3  \\ \hline
\end{tabular}}
\caption{Translation performance of various families of
LLMs with a similar size on human-collected unfaithful translation test sets.}
\label{model_size_unfaithful}
\end{table*}

\end{document}